\documentclass[10pt,twocolumn,letterpaper]{article}

\usepackage{cvpr}              % To produce the CAMERA-READY version
\usepackage{diagbox}

\usepackage{subcaption}
\usepackage{pgfplots}
\pgfplotsset{width=7cm,compat=1.16}
\usepgfplotslibrary{external}
\definecolor{cvprblue}{rgb}{0.21,0.49,0.74}
\usepackage[pagebackref,breaklinks,colorlinks,allcolors=cvprblue]{hyperref}

\def\paperID{16009} % *** Enter the Paper ID here
\def\confName{CVPR}
\def\confYear{2025}

\title{Do ImageNet-trained models learn shortcuts?  \\ The impact of frequency shortcuts on generalization    }

\author{Shunxin Wang \quad Raymond Veldhuis \quad Nicola Strisciuglio\\
University of Twente\\
{\tt\small \{s.wang-2, r.n.j.veldhuis, n.strisciuglio\}@utwente.nl}
}

\begin{document}
\maketitle
\begin{abstract}
Frequency shortcuts refer to specific frequency patterns that models heavily rely on for correct classification. Previous studies have shown that models trained on small image datasets often exploit such shortcuts, potentially impairing their generalization performance. However,  existing methods for identifying frequency shortcuts require  expensive computations and become impractical for analyzing models trained on large datasets.  In this work, we propose the first approach to more efficiently analyze frequency shortcuts at a large scale.  We show that both CNN and transformer models learn frequency shortcuts on ImageNet. We also expose that frequency shortcut solutions can yield good performance on out-of-distribution (OOD) test sets which largely retain texture information. However, these shortcuts, mostly aligned with texture patterns, hinder model generalization on rendition-based OOD test sets. These observations suggest that current OOD evaluations often overlook the impact of frequency shortcuts on model generalization. Future benchmarks could thus benefit from explicitly assessing and accounting for these shortcuts to build models that generalize across a broader range of OOD scenarios. 
Codes are available at \url{https://github.com/nis-research/hfss}.
\end{abstract}    
\section{Introduction}
\label{sec:intro}
%Shortcut solutions based on 
Superficial correlations between data and ground truth~\cite{Geirhos_2020,06406} can be learned by models to minimize training objectives with the least effort~\cite{NEURIPS2020_6cfe0e61}. This learning behavior is called shortcut learning, which either harms the generalization performance of models or gives an illusion of good generalization abilities when learned shortcuts are present in OOD test sets~\cite{Wang_2023_ICCV}.  Artifacts and visual cues that cause shortcut learning are recognizable by visual inspection of images, and their impact on the training of models can be mitigated through data selection or augmentation~\cite{diagnostics12010040,06406,lapuschkin2019unmasking}, i.e. counteracting the spurious correlations between data and ground truth. 

However, there exist shortcuts in the Fourier domain, which are implicitly embedded in image data characteristics and not easily detectable by visual inspection~\cite{Wang_2023_ICCV,wang2022frequency}. Such shortcut solutions consist of small frequency subsets  that are easy-to-learn and sufficient for models to achieve high classification rate. Typically, they correspond to simple features like textures, shapes or colors in the spatial domain. 
Wang, et al.~\cite{Wang_2023_ICCV} identified frequency shortcuts by retaining relevant frequencies to classification of a certain class. The relevance of an individual frequency was measured by the loss value of the model tested on images of the class with that frequency removed. Thus, their approach requires computational time increasing proportionally to the number of classes in a dataset and the image resolution. 
% Measuring the relevance of individual frequency for classificatin neglects correlations among frequencies.
Yet, frequency shortcuts usually consist of multiple frequencies. The method in~\cite{Wang_2023_ICCV} might overlook some shortcuts as individual frequency relevance to classification  neglects joint contribution of frequencies. Furthermore, the computational burden limits its applicability for analyzing shortcut learning behavior of models trained on large datasets (e.g. ImageNet) with hundreds or thousands of classes. 
\begin{table}
\footnotesize
    \centering
    \caption{Our method needs less computation times compared to~\cite{Wang_2023_ICCV} (using ResNet18 on an NVIDIA A40 GPU). Computational time of~\cite{Wang_2023_ICCV} on ImageNet-1k is estimated from their ImageNet-10 experiments, considering that it increases proportionally to the number of classes.  }
    \begin{tabular}{ccc}
    \toprule
        Dataset &   Time (h)~\cite{Wang_2023_ICCV} &  Time (h) (Ours) \\
         \midrule
        CIFAR-10 & 7.5 & \textbf{0.5} \\
          % \midrule
          ImageNet-1k & 8500 (354 days) & \textbf{ 220 (9.2 days)} \\  %ViT-B: IN-1k, 1357 (56.5 days)
        \bottomrule
    \end{tabular}
    \vspace{-2em}
    \label{tab:comparsion_time}
\end{table}
Several studies have shown that ImageNet-trained models are biased towards textures~\cite{geirhos2018imagenettrained,Gavrikov_2024_CVPR}. However, there is no concrete evidence of what causes this phenomenon, although~\cite{Wang_2023_ICCV} largely attributes it to frequency shortcut learning.  

In this work, we propose the first method that enables the uncovering of frequency shortcuts learned by models trained on large-scale datasets (e.g. ImageNet-1k), for analyzing learning behavior and explaining model generalization performance in different OOD scenarios.  Our method improves computational efficiency  as it applies parallel class-wise loss computation and hierarchical search in the Fourier domain (see~\cref{tab:comparsion_time}), compared to~\cite{Wang_2023_ICCV}.  Moreover, our method considers the joint contribution of frequencies to classification, thus being more effective in finding shortcuts.
Our contributions are:
\begin{enumerate}
    \item We develop a hierarchical frequency shortcut search (HFSS) method, which enables the analysis of shortcut learning in the Fourier domain on large datasets with varying class counts. We reduce computational time and improve the effectiveness at identifying shortcuts.
    \item We discover that ImageNet-trained models (both CNN and transformer architectures) are subject to learn frequency shortcuts. Different from shortcuts formed by visual cues~\cite{Geirhos_2020}, frequency shortcuts (easy-to-learn features) lead to good performance on both in-distribution (ID) and OOD tests if they do not block models from learning other semantics (difficult features). 
    \item Our HFSS enables a more comprehensive assessment  of model generalizability  by analyzing OOD data characteristics, specifically the presence of shortcuts.
    % , a factor often overlooked in existing OOD evaluation frameworks.   
    In existing evaluation frameworks, frequency shortcuts do not always impair OOD generalization performance. This emphasizes the importance of considering the role of shortcuts when designing future OOD  evaluation benchmarks. % for OOD generalization.  
    % We observe that whether frequency shortcuts impair or yield good performance on OOD test sets depends on  the availability of specific texture characteristics (corresponding to most frequency shortcuts) in the OOD data.
\end{enumerate}

% \noindent  Within current OOD evaluation benchmarks, which measure model performance on additional data without considering data characteristics, frequency shortcuts are not found to be necessarily harmful to OOD generalization. This emphasizes the need to account for the impact of shortcuts when designing future  evaluation benchmarks for OOD generalization.  

\section{Related works}
\label{sec:relatedwork}
\paragraph{Shortcut learning.}
Models can learn shortcut solutions based on superficial correlations between data and ground truth~\cite{Geirhos_2020,gu2019understanding} to optimize training objectives with the least effort. This learning behavior is due to simplicity-bias~\cite{NEURIPS2020_6cfe0e61}, which is caused by inductive biases provided by gradient descent or components like ReLUs~\cite{Teney_2024_CVPR}. Shortcuts can be visual cues in the data like source tags and artificial markers~\cite{lapuschkin2019unmasking, diagnostics12010040}.  For instance, one fifth of the horse images from Pascal VOC dataset were found to contain a source tag, on which models rely as a discriminant feature to recognize horses~\cite{lapuschkin2019unmasking}. 
Next to these visual cues, vision models are also subject to shortcuts implicitly existing in the frequency domain, which manifest as  small sets of frequencies contributing significantly to image classification performance~\cite{Wang_2023_ICCV,wang2022frequency}.  The study in~\cite{Wang_2023_ICCV} explains why models exhibit a textures-bias for classification~\cite{geirhos2018imagenettrained} from a frequency shortcut perspective, but does not provide further investigation on how frequency shortcuts affect the generalization and robustness performance of ImageNet-trained models, due to computational burden. 

Relying on simple shortcut solutions could harm the generalization and robustness performance of models, as shortcut learning appearing in early training might block models from learning other semantics related to the tasks at hand~\cite{cao2020understanding,CiCP-28-1746,14313,Wang_2023_ICCV}. However, such reliance could also creates a false impression of good generalization when shortcuts  are present in OOD test sets ~\cite{pmlr-v119-bahng20a,Wang_2023_ICCV,lapuschkin2019unmasking}. 

\paragraph{Shortcut identification.}
Uncovering shortcuts helps understanding the generalizability of models on different out-of-distribution (OOD) data, explaining why models fail to generalize to or perform well on OOD data. It can provide shortcut-related prior knowledge to develop techniques that improve model generalization and robustness performance based on shortcut mitigation~\cite{Wang_2023_ICCVDFM,diagnostics12010040}.  However, common approaches are limited to identifying visually inspectable shortcuts such as text, watermark and color patches~\cite{lapuschkin2019unmasking,diagnostics12010040,08822}, using e.g. saliency maps~\cite{8237336}. 
Rather than directly identifying shortcuts in the data,  in \cite{06406,boland2024there}  shortcut features present in individual images were  quantified  by assessing how difficult they were  for models to learn.
The authors in~\cite{NEURIPS2022_baaa7b5b,pezeshki2021gradient,11230} uncovered shortcut features learned within representation space.
The authors in~\cite{marconato2023not} investigated shortcut learning by analyzing the relationship between  semantic concepts using a knowledge graph, sharing a similar idea to~\cite{NEURIPS2022_536d6438}. 
Despite visual shortcut cues, there are non-observable shortcuts in the frequency domain, which are embedded in data characteristics.
The work of~\cite{Wang_2023_ICCV} identified relevant frequencies to classification, which potentially contain shortcut information. As this technique measures the relevance of one frequency at a time, it requires computational costs increasing proportionally to the number of classes and image resolution. Thus, limited attention has been given to uncovering frequency shortcuts in large datasets, with existing analyses primarily focused on datasets containing a small number of classes~\cite{Wang_2023_ICCV,boland2024there,NEURIPS2022_536d6438,wang2022frequency}.

Existing methods to identify shortcuts are limited to either manual inspection~\cite{lapuschkin2019unmasking} or involve a time-consuming algorithm~\cite{Wang_2023_ICCV,wang2022frequency}. Methods quantifying shortcut information or uncovering shortcut features primarily aim at mitigating shortcut learning, without explaining how shortcuts in the data impact model generalization performance. We are the first to enable frequency shortcut analysis of models trained on large-scale datasets and link them to generalization performance in different OOD settings.

\begin{figure*}[!t]
    \centering
   \includegraphics[width=0.9\textwidth]{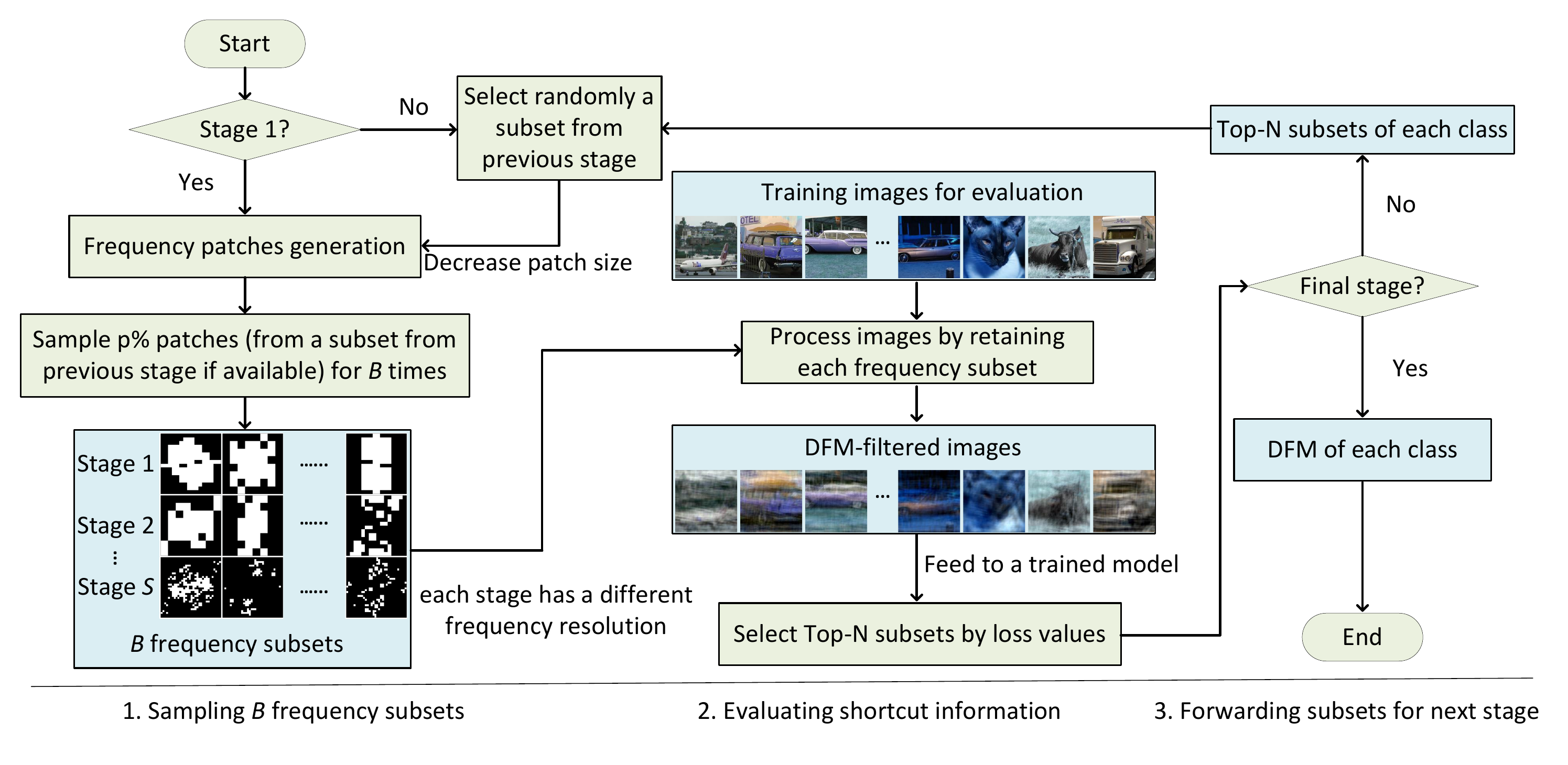}
   \vspace{-1em}
    \caption{Scheme of HFSS. Starting from stage 2, we sample frequency patches from  a random frequency subset searched in previous stage. This confines the size of search space. The white patches in the binary masks indicate sampled frequency patches.    }
    \vspace{-1em}
    \label{fig:scheme_general}
\end{figure*}

\paragraph{OOD evaluation.} 
The  generalization and robustness performance of vision models are usually evaluated using extra data that is considered OOD, e.g. data collected at different time points~\cite{pmlr-v97-recht19a},  with different styles or renditions~\cite{Hendrycks_2021_ICCV,geirhos2018imagenettrained,wang2019learning}, with synthetic corruption effects~\cite{hendrycks2018benchmarking,kar20223d,NEURIPS2021_1d497805,wang2023larger}, and with adversarial noise~\cite{NEURIPS2023_d9888cc7,GUO2023109308,croce2021robustbench,8835364}. However, such  benchmarks do not consider the impact of shortcuts learned by models and present in OOD test sets. This might ignore critical factors related to the generalization and robustness capabilities of models. The work in~\cite{Gavrikov_2024_CVPR} explores model generalization from the  perspectives of biases, e.g. texture, shape and spectral biases. 
 In this work, we investigate  the impact of frequency shortcuts on model generalization abilities, establishing connections among these different biases. Our work provides insights into when frequency shortcuts yield good performance   or are harmful to model generalization and robustness performance.
 
 % discover that frequency shortcuts learned by models  benefit model generalization performance if they present in the OOD data and do not block the models from learning other related semantics during training. They impair generalization performance, in the case when they are absent from the OOD data, e.g.  ImageNet-trained models tested on images with missing texture information, which correspond to most frequency shortcuts.

\section{Method}
\label{sec:method}
We propose a method to identify (non-visible) frequency shortcuts  linked to intrinsic data characteristics rather than visual cues as in~\cite{diagnostics12010040}. We reduce significantly computational demands compared to~\cite{Wang_2023_ICCV}.  Our method is the first solution for shortcut identification that enables the analysis of models trained on datasets with a large number of samples and classes. In the following sections, we detail our measurement of shortcut learning in models and examine the impact of shortcuts on generalization.

\subsection{Identifying frequency shortcuts}
%To identify frequency shortcuts learned by models trained on large datasets with thousands of classes, 
We propose a method called hierarchical frequency shortcut search (HFSS), which exploits hierarchical search of frequency subsets in the image Fourier spectrum. This reduces computational time compared to exhaustive search strategies, e.g.~\cite{Wang_2023_ICCV,Wang_2023_ICCVDFM}.  The search is separated into several stages, each stage gradually  narrowing down the spectrum search space in a coarse-to-fine manner and detecting frequency subsets that models strongly rely on for classification. The scheme of HFSS is shown in~\cref{fig:scheme_general}. 

\paragraph{Hierarchical search for frequency shortcuts.}
In order to discover the frequency subsets that models rely heavily on for correct classification of each class, we sample different frequency combinations that potentially contain frequency shortcuts. We use random sampling as it allows to easily consider joint contribution of frequencies to classification,  improving the effectiveness of frequency shortcut identification. Random sampling has been shown to contribute to stable search results in data augmentation and reinforcement learning~\cite{lim2019fast,müller2021trivialaugment,pmlr-v70-osband17a}, while requiring lower computations w.r.t. optimization-based methods. 

 \begin{figure}[!t]
    \centering

    \includegraphics[width=0.25\textwidth]{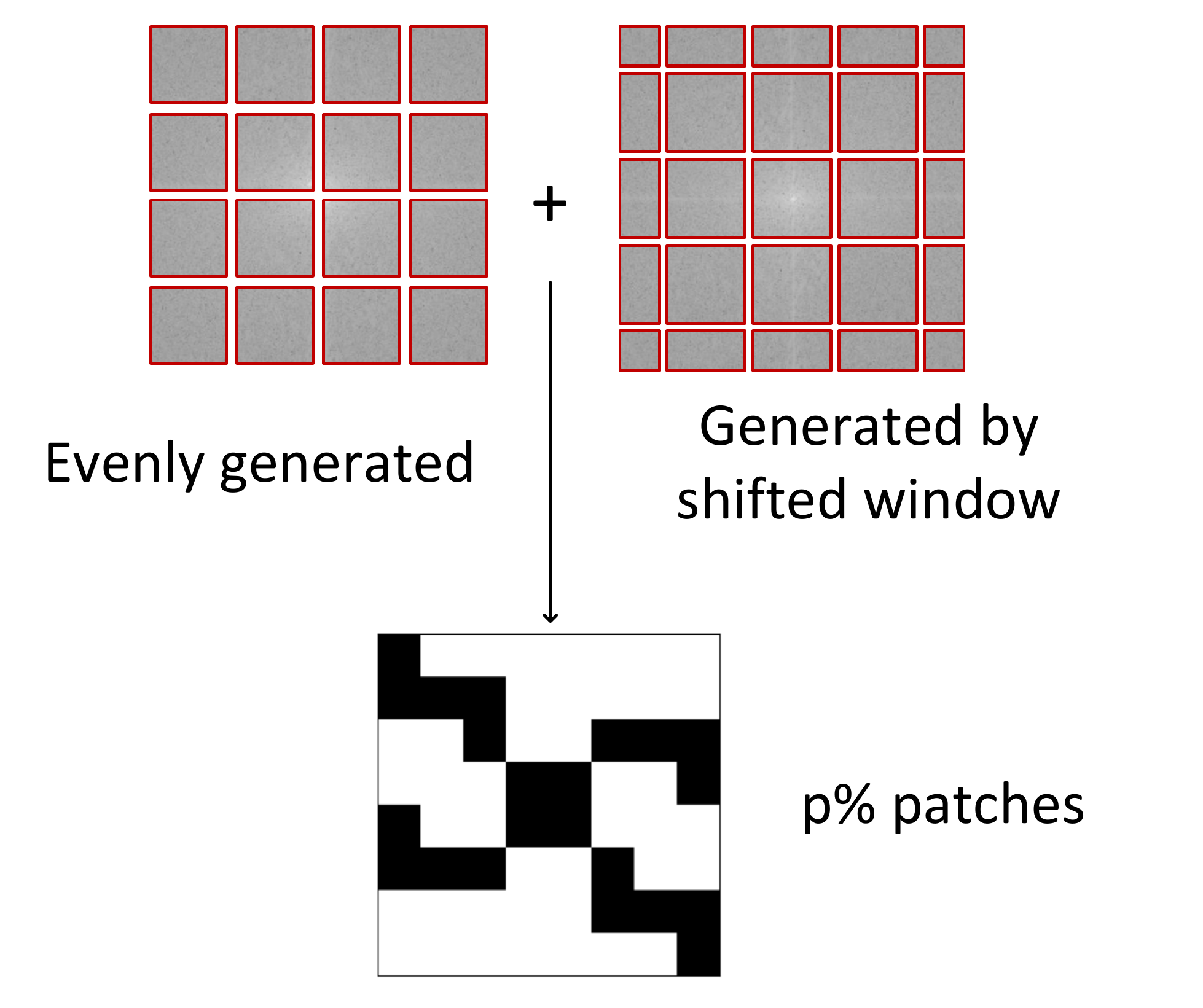} 
    \vspace{-1em}
    \caption{The frequency spectrum is separated into patches, with $p\%$ sampled for shortcut evaluation. }
    \vspace{-1.8em}\label{fig:patch_shiftwindow}
\end{figure}

There are numerous combinations of frequencies by sampling frequency components directly. Instead of doing an exhaustive search, we apply hierarchical search in the Fourier spectrum: we divide the Fourier spectrum into frequency patches and sample these frequency patches to generate candidates of frequency subsets that contain shortcut information, as illustrated in~\cref{fig:patch_shiftwindow}.  We exploit overlapping shifted windows in vertical and horizontal directions~\cite{liu2021swin} to prepare frequency patches and thus avoid border effects, in addition to evenly separating the Fourier spectrum. 
% During hierarchical search, we gradually decrease the patch size and generate candidates from the smaller frequency patches, thus improving the frequency resolution of identified frequency subsets. 
%
 HFSS consists of multiple search stages, the number of which depends on image resolution and desired frequency resolution of the identified frequency subsets. Lower image or frequency resolution requires fewer search stages.

 In summary, the overall process of HFSS is as follows. 
 Each stage of HFSS consists of three steps, namely (1) sampling  $B$ frequency subsets which contain $p\%$ frequency patches, (2) evaluating shortcut information, and (3) forwarding subsets for next stage. 
 At the initial search stage, we sample frequency patches with a large size and evaluate their shortcut information. 
 We evaluate the frequency subsets according to their contributions to classification: we process a subset of training images by retaining a frequency subset and measure the class-wise loss of a model. 
Frequency subsets that contribute to a lower loss value are stronger candidates to indicate frequency shortcuts as they are sufficient for the model to achieve a high prediction score.  
We rank frequency subsets by the loss values and use the top-$N$ masks of each class in the next search stage.  

 Starting from the second search stage, we sample frequency patches from one of the top-$N$ frequency subsets in the previous stage,  with a smaller patch size. This confines the search space.
 In the final search stage, the top-$1$ frequency subset of each class is considered dominating the classification of the class concerned. Same as~\cite{Wang_2023_ICCV}, we use binary masks to represent these frequency subsets, called \textbf{D}ominant \textbf{F}requency \textbf{M}aps ($\mathrm{DFMs}$).

 The sampling percentage $p\%$ of frequency patches, the number of sampled frequency subsets $B$ and the frequency resolution of  DFMs are hyperparameters of the search algorithm (details are in the supplementary material).

\subsection{Measuring the degree of shortcut learning}
Analyzing models learning behavior in a class-wise manner, as done in~\cite{Wang_2023_ICCV}, is labor-intensive if there are hundreds or thousands of classes.  To provide a broad overview of model learning behavior on training data, we categorize all classes in a dataset into two groups: (1) classes subject to shortcuts and (2) non-shortcut classes. A class is considered subject to shortcuts if its true positive rate (TPR) surpasses a given threshold when the model is tested on  images filtered to retain only dominant frequencies of the class concerned. We use TPR, as a high TPR indicates that the frequency subset is sufficient to achieve a high classification rate, thereby acting as a shortcut. The frequency-filtered images are referred to as \textbf{DFM-filtered images}.

\begin{figure}
    \centering
    \includegraphics[width=0.65\linewidth]{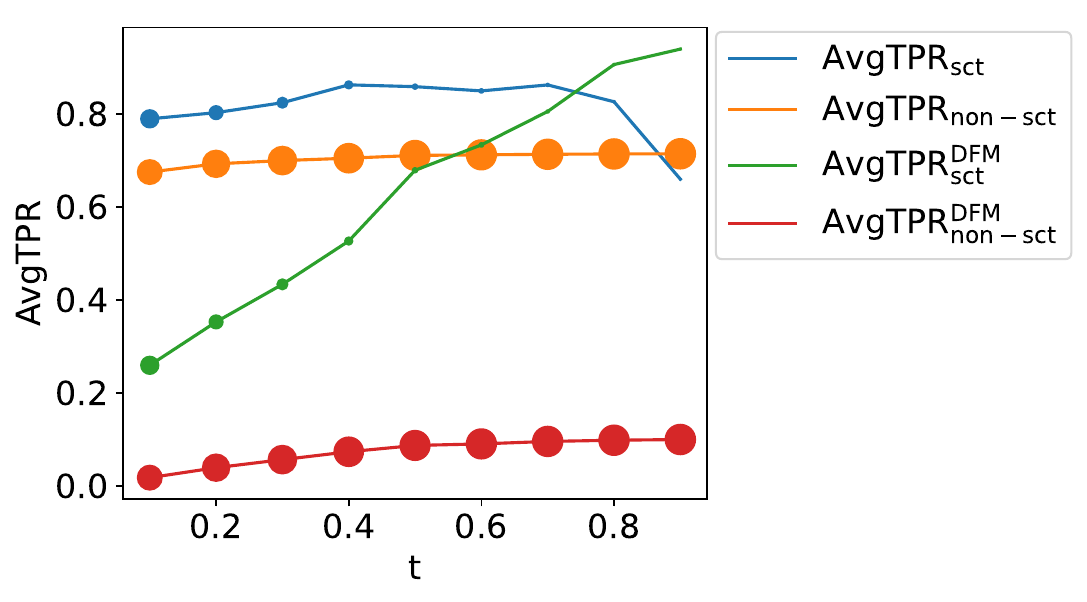}
\vspace{-1em}
    \caption{ResNet18 tested on original test images and DFM-filtered images of ImageNet-1k. Blue line shows the average TPR of classes subject to shortcuts on the original test images, and the orange line shows the results of non-shortcut classes, at different threshold $t$.  The green and red lines correspond to results tested on DFM-filtered images.  Lower $t$ indicates weak shortcuts and higher $t$ signifies stronger ones. The size of each point reflects the number of classes. The larger the size, the more classes included.  }
    \vspace{-1em}
    \label{fig:sample_eval}
\end{figure}

For each class, we test the model on both the original test images and DFM-filtered images, computing TPR for both. We denote the TPR on the original images of class $c_i$ as $\mathrm{TPR}_{c_i}$ and the TPR on the DFM-filtered images as $\mathrm{TPR}^{DFM}_{c_i}$.  If $\mathrm{TPR}^{DFM}_{c_i}{>} t$ (where $t{\in}[0,1]$ is a predefined threshold), class $c_i$ is considered subject to shortcuts; otherwise, it is considered as one of the non-shortcut classes.
We compute the average $\mathrm{TPR}$ values of both groups,  shortcut and non-shortcut classes,  at different thresholds $t$. We note them as $\mathrm{AvgTPR_{sct}@t}$, $\mathrm{AvgTPR^{DFM}_{sct}@t}$ for the shortcut classes, and  $\mathrm{AvgTPR_{non-sct}@t}$  and $\mathrm{AvgTPR^{DFM}_{non-sct}@t}$ for the non-shortcut classes. A higher threshold $t$ corresponds  to stronger shortcuts, as dominant frequencies alone are sufficient for accurate classification.
An example of the results of these metrics at different threshold values are given in~\cref{fig:sample_eval}, where the size of each point reflects the number of classes in the two groups. Larger points indicate a higher number of classes, which suggests that the model has a stronger tendency to shortcut learning.

We do not average  $\mathrm{TPR^{DFM}}$ across all classes because its value for many non-shortcut classes is close to zero (specifically for large datasets). This can result in a close-to-zero $\mathrm{AvgTPR^{DFM}}$ which does not provide useful information for further analysis. 

\subsection{Impact of shortcut on generalization}
To assess how frequency shortcuts impact the generalization and  robustness performance of models trained on ImageNet-1k, we evaluate their performance on several datasets. 
We perform ID tests on ImageNet-1k (IN-1k)~\cite{5206848} and ImageNet-v2 (IN-v2)~\cite{pmlr-v97-recht19a} test sets. They are considered in-distribution as they are collected using the same protocol as the training set. 
Furthermore, we use ImageNet-C (IN-C)~\cite{hendrycks2018benchmarking} to benchmark the robustness of models against appearance image corruptions, e.g. noise,  blur and weather changes. 
To evaluate the generalizability of models to images with different renditions, we use ImageNet-Renditions (IN-R)~\cite{Hendrycks_2021_ICCV} and ImageNet-Sketch (IN-S)~\cite{wang2019learning}. IN-R contains renditions like cartoons, paintings, art and toys of 200 classes from IN-1k. 
IN-S has the same number of classes as IN-1k, containing images of hand-drawn sketches. 
These datasets contain visual renditions, which serve to evaluate how the reliance on texture cues (corresponding to most frequency shortcuts) impacts on generalization. 
To measure the adversarial robustness of models, we apply the fast gradient sign method (FGSM) attacks~\cite{goodfellow2015explainingharnessingadversarialexamples} to the validation set of IN-1k, with $L_\infty = 4/255$. We calculate the average TPR of shortcut and non-shortcut classes on OOD data, that we compare to the results on ID data.

\section{Experiments}
\label{sec:exp}

% In HFSS, the number of sampled frequency subsets $B$ can be configured as hyperparameters. 
We start with investigating the trade-off between efficiency (i.e. required computational time) and effectiveness (i.e. capability of finding shortcuts),  varying the configuration of the number of sampled frequency subsets at each stage, noted as $B_s$ ($s$ is the index of stage).  
Then, we analyze frequency shortcut learning of models trained on IN-1k, relating it to performance results under different OOD scenarios.

\paragraph{Setup.} 
We use CIFAR-10 (C-10)~\cite{Krizhevsky_1970} to explore the configuration of $B$, and configure HFSS  with four search stages,  with patch size of $8{\times}8$, $4{\times}4$, $2{\times}2$ and $1{\times}1$, respectively.  
We then perform larger-scale shortcut analysis using IN-1k: we use six search stages, with patch size of $56{\times}56$, $28{\times}28$, $14{\times}14$, $8{\times}8$, $4{\times}4$ and $2{\times}2$. The patch (frequency) resolution in the final stage is the same as that in~\cite{Wang_2023_ICCV}. At each stage, we sample $p{=}60\%$ frequency patches to form frequency subsets. It results in  $\mathrm{DFMs}$ containing about $5\%$ frequencies of the whole spectrum of ImageNet images and about $15\%$ frequencies of CIFAR images. %about 13\%
The number of training images sampled for shortcut information evaluation is the same as that of their corresponding test set. 
We evaluate the generalization and robustness performance of ImageNet-trained models on IN-v2~\cite{pmlr-v97-recht19a}, IN-C~\cite{hendrycks2018benchmarking}, IN-R~\cite{Hendrycks_2021_ICCV},  IN-S~\cite{wang2019learning} and  using FGSM~\cite{goodfellow2015explainingharnessingadversarialexamples} attacks. 
We also carry out experiments on ImageNet-10 (IN-10)~\cite{huang2021unlearnable} with a ResNet18 model to compare HFSS with single-frequency removal-based method in~\cite{Wang_2023_ICCV}. Following their setups, we use ImageNet-SCT (IN-SCT) for OOD evaluation. Our configuration of HFSS enables direct comparison with the results reported in~\cite{Wang_2023_ICCV} as  $\mathrm{DFMs}$ contain around $5\%$ frequencies of whole spectrum. 
We report training configurations and additional results in the supplementary material.

\subsection{Configuration of HFSS}
\label{sec:opti}
\paragraph{Sampled frequency subsets.} At stage $s$, we sample  $B_s$ frequency subsets, such that HFSS can explore different combinations of frequency patterns, validating their contained shortcut information. However, sampling an excessive number of subsets  increases the required search time, as more subsets are  for evaluating their contributions to classification results. To explore the impact of the number of sampling operation $B_s$ on shortcut identification, we perform experiments on C-10 using a ResNet18 backbone, setting the number of sampled frequency subsets in each stage to be as large as possible under limited computational time. We hypothesize that HFSS obtains stable results given enough candidate frequency subsets. 
We set  $B_1$, $B_2$, $B_3$, and $B_4$ to be 1000, 2000, 4000 and 8000 respectively, which we note as configuration CF-1. The increasing number of candidate frequency subsets at consecutive stages is determined by the increase in possible frequency patch combinations as patch size decreases.

\begin{figure}
\includegraphics[width=\linewidth]{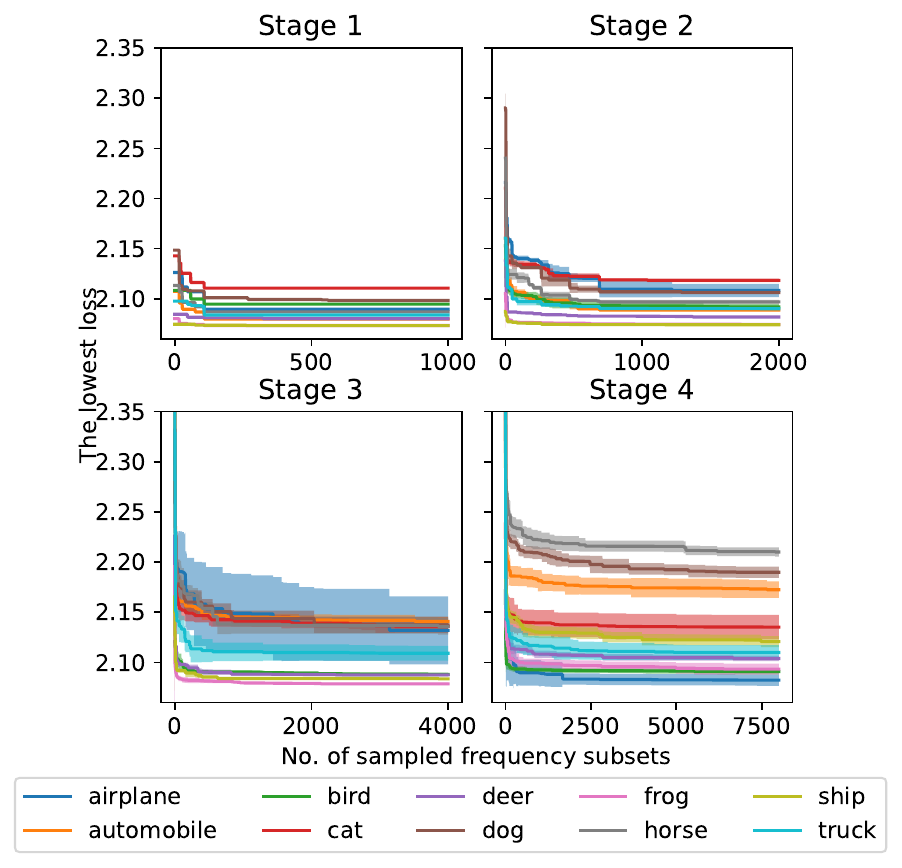}
\vspace{-2em}
  \caption{The best class-wise loss vs. the number of sampled frequency subsets at each stage.}
  \vspace{-1.5em}\label{fig:lossvsnosamplessrn18}

\end{figure}
We run the  configuration CF-1 five times and note the stability of our HFSS search algorithms. We track the best (lowest) loss values achieved for each class when the model is tested on DFM-filtered images.  
In~\cref{fig:lossvsnosamplessrn18}, we report the tracking statistics with average loss values (lines) and standard deviations (shadows) over the five trials for each class, which are further used as a starting point to optimize the configuration of HFSS. As the number of sampled frequency subsets increases, the best loss for each class converges, indicating \textbf{the stability of HFSS in identifying similar shortcuts learned by models when a sufficient amount of candidate frequency subsets are sampled for validating shortcut information}.
The standard deviation recorded at different stages is relatively small compared to the range of best loss. For class \textit{airplane} we observe an outlier outcome at stage $3$. In the supplementary material, we present visualization of shortcut cues (DFM-filtered images) computed across the five trials. These visualizations show that the identified shortcut patterns differ in the orientations of strip-like features, corresponding to line features in the image of \textit{airplane}. This suggests that frequency shortcuts do not arise from fixed frequency subsets; rather, spatial features they correspond to may be visually similar but composed of different frequencies, determining an higher standard deviation in~\cref{fig:lossvsnosamplessrn18}. %This could attribute to the relatively high standard deviations.
% conjecture that the model exploits different shortcut cues, as observed from the visualization of the shortcuts uncovered over the five trials (see supplementary material).

\paragraph{Efficiency vs. effectiveness.}
\begin{figure}
   \centering
    \includegraphics[width = 0.28\textwidth]{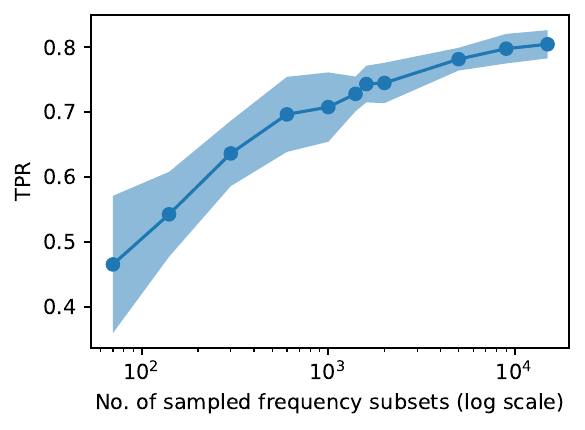} 
    \vspace{-1em}
    \caption{Average TPR vs. search time,  where search time increases proportionally with the number of sampled frequency subsets. }
    \vspace{-1.5em}
   \label{fig:perfvstime}
\end{figure} 
Based on the results in~\cref{fig:lossvsnosamplessrn18}, we optimize CF-1 for efficiency and effectiveness in shortcut identification. The required search time scales up proportionally to $B_s$, i.e. the more candidate frequency subsets are sampled, the more time is needed to verify their relevance to classification. 
We perform experiments using different $B_s$, investigating how the size of the search space affects the identification of shortcuts. Based on the tracking statistics in~\cref{fig:lossvsnosamplessrn18}, which shows that increasing $B_s$ beyond a certain point does not decrease loss significantly, we adjust $B_1$ to $200$, $B_2$ to $800$ and $B_4$ to $4000$ (CF-2.1).  We further reduce $B_3$ and $B_4$ to $2000$ (CF-2.2). We also design eight other configurations (see details in the supplementary material), with CF-2.10 being the most efficient as  it samples the lowest number of candidate frequency subsets.

To compare the performance of HFSS under different configurations, we compute the average TPR over all classes when tested on DFM-filtered images.  This indicates the general relevance of the searched frequency subsets to classification. A higher average TPR indicate higher relevance, thus the frequency subsets contain more frequency shortcut information. We run each configuration five times, computing the mean and standard deviations of average TPRs over the five trials. 
We report the results in~\cref{fig:perfvstime} where the left-most point corresponds to average TPR  of CF-2.10 and the right-most point corresponds to that of CF-1. We observe that as more frequency subsets are sampled, average TPR increases and saturates. This demonstrates that sampling more candidate frequency subsets does not significantly improve the performance of HFSS, while requiring more computations. 

\begin{figure}
    \centering
    \includegraphics[width=0.55\linewidth]{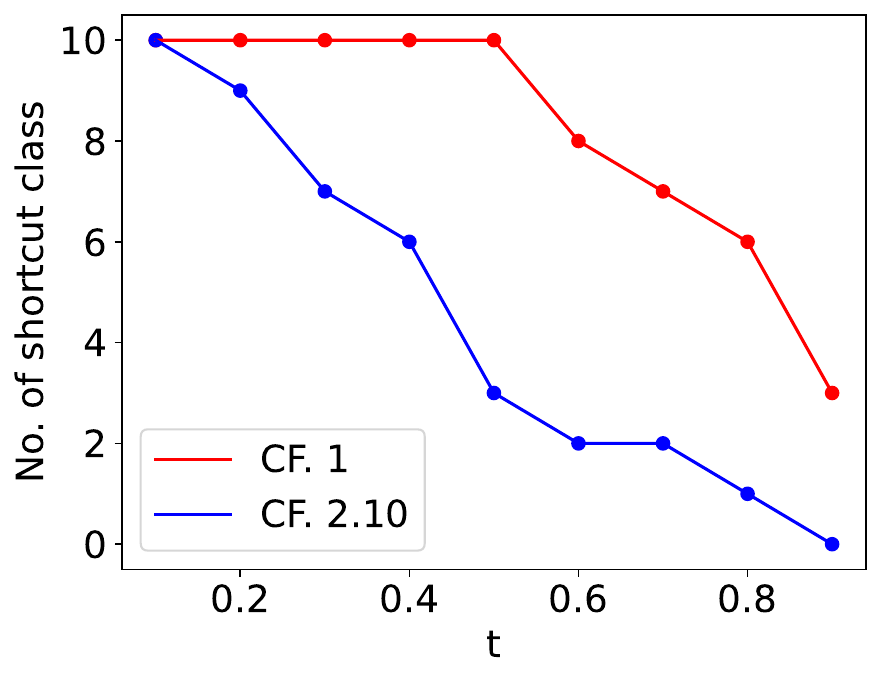}
    \vspace{-1em}
    \caption{The number of shortcut classes given different threshold $t$ search by CF-1 and CF-2.10. Using CF-2.10 uncovers most shortcuts identified by using CF-1 at low thresholds, with computational time reduced to a factor of around 200.  }
    \vspace{-1em}
    \label{fig:cifarconf1_21}
\end{figure}

 We compare CF-1 and CF-2.10 in~\cref{fig:cifarconf1_21} based on the number of classes with identified shortcuts at different thresholds. We observe that CF-2.10 uncovers most of the shortcuts identified by using the more complex CF-1 at low threshold level. Although CF-2.10 misses stronger shortcuts (at higher threshold levels),  it  achieves a ${\sim}200{\times}$ reduction in computational time compared to CF-1. 
A more time-efficient configuration is less effective at finding shortcuts than the more complex one,  but it still manages to identify shortcuts at low thresholds.  

The observations made on C-10 allow to estimate initial configurations for analysis on larger-scale datasets. We track the loss statistics of HFSS applied to ImageNet  (provided in the supplementary material) and choose the value $B_s$ for   ImageNet experiments in~\cref{sec42} as the one where the loss values show no  significant decrease. Given the computational intensity of identifying shortcuts in large-scale datasets, our work on IN-1k primarily aims at uncovering shortcuts with affordable computational efforts.

\begin{figure*}
    \centering
    \includegraphics[width=0.9\linewidth]{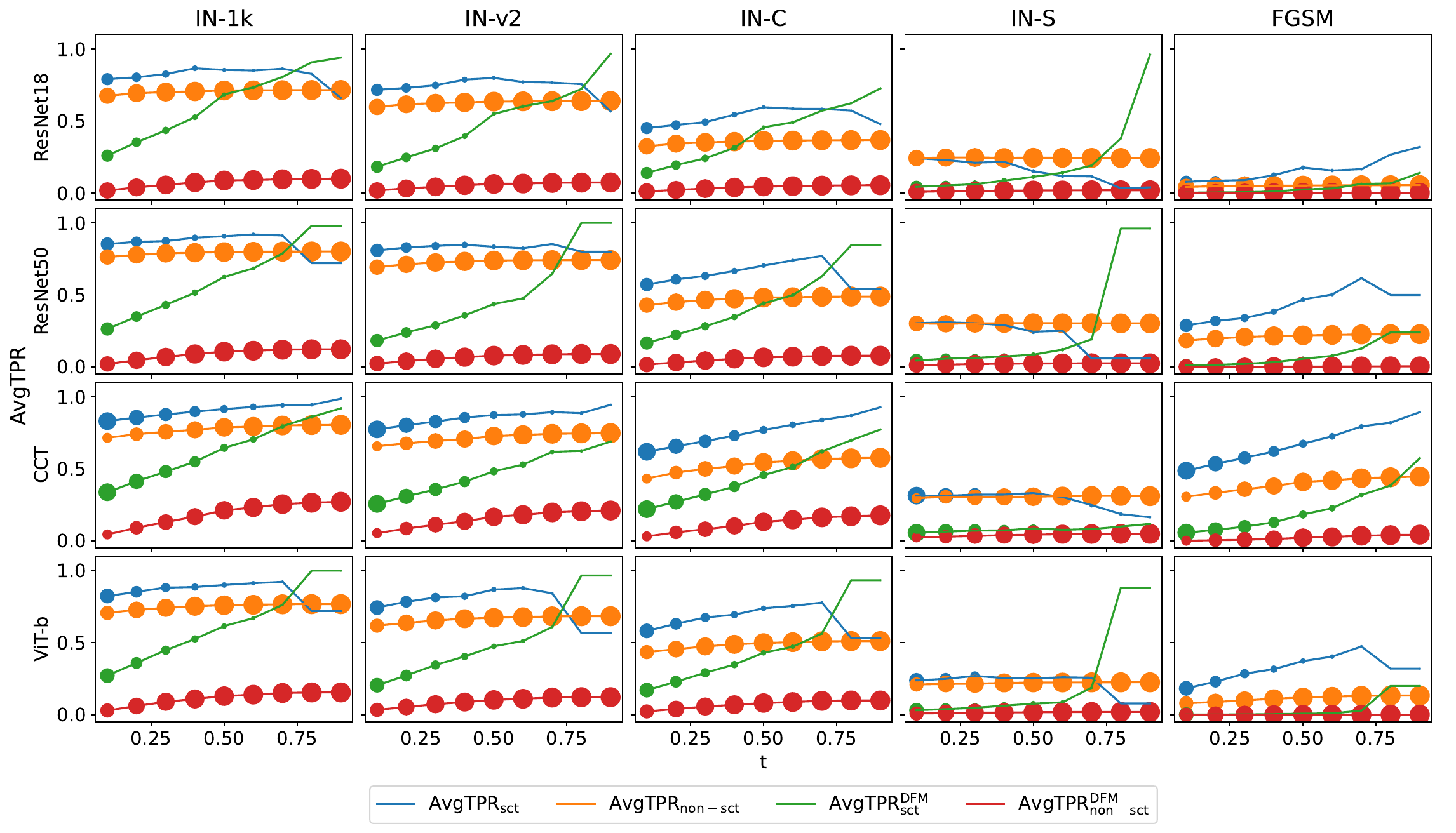} 
    \vspace{-1em}
    \caption{Average TPR of shortcut and non-shortcut classes given different thresholds on datasets with 1000 classes.  Models generally perform better on images of shortcut classes than non-shortcut classes. This does not hold for results on IN-S, which lacks preserved texture information. }
    \vspace{-1em}
   \label{fig:in1k_threshold}
\end{figure*}

\begin{figure}
    \centering
    \includegraphics[width=0.8\linewidth]{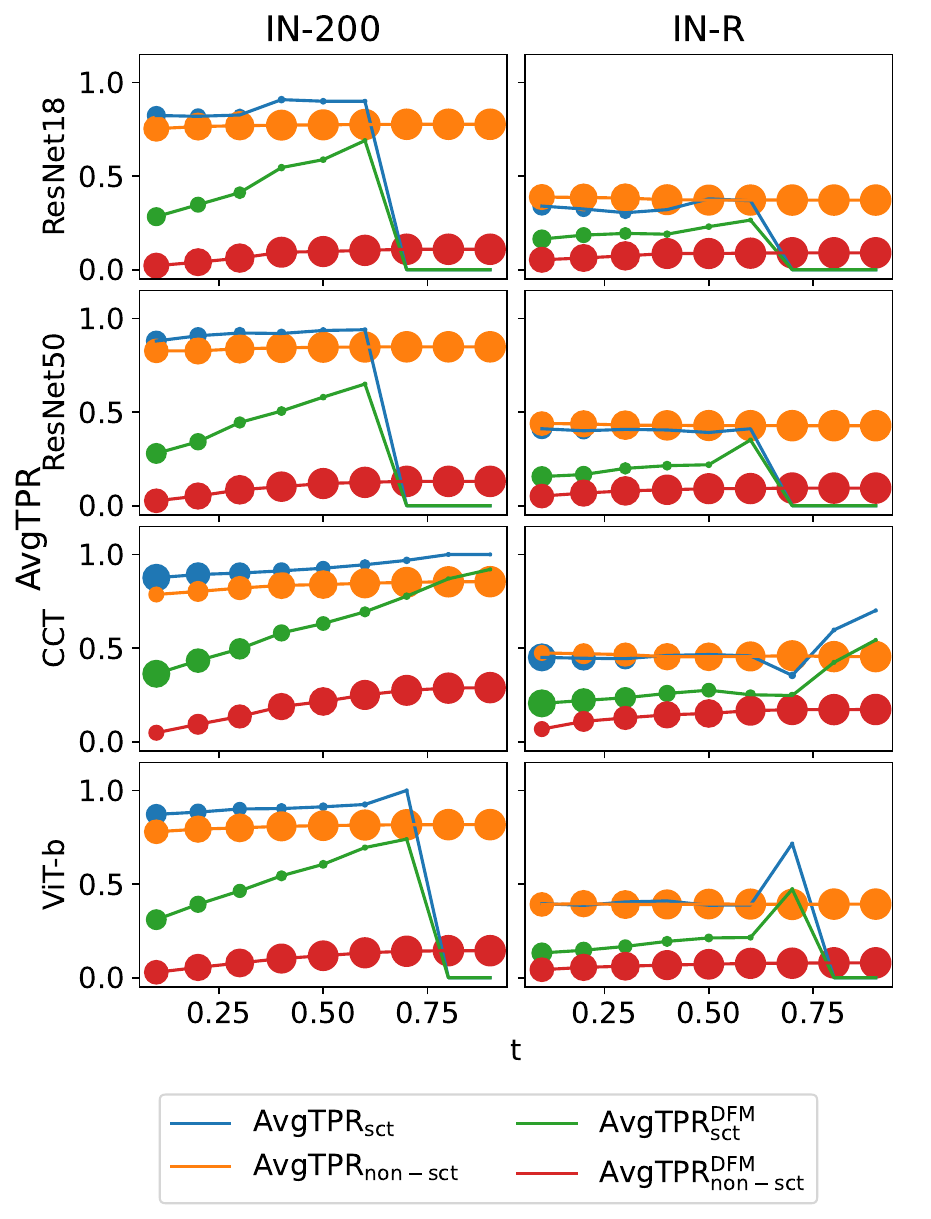}
    \vspace{-1em}
    \caption{Average TPR of shortcut and non-shortcut classes at different thresholds on datasets with 200 classes.  Models perform better on shortcut classes than non-shortcut classes in IN-200. IN-R lacks preserved texture information and thus the models have worse performance on shortcut classes than non-shortcut classes. }
    \vspace{-2em}
     \label{fig:inR_threshold}
\end{figure}

\subsection{Shortcut identification on IN-1k}
\label{sec42}
We apply HFSS to examine frequency shortcuts learned by ImageNet models and how they impact model generalization and robustness.  The search configurations for ImageNet models, aimed at time-efficiency and effectiveness at finding shortcuts, are in the supplementary material.

\paragraph{Results on IN-1k, IN-v2 and  IN-C. }
In~\cref{fig:in1k_threshold,fig:inR_threshold}, we report the results of the analysis of impact of shortcuts on model performance.
In general, models perform better predictions on images of classes affected by shortcut learning (blue lines) compared to non-shortcut classes (orange lines) in both ID and corruption tests. Models subject to frequency shortcuts yield good performance in the robustness and generalization tests when textures information is largely preserved. For ResNet and ViT models, when $t{\geq}0.8$,  the $\mathrm{AvgTPR^{DFM}_{sct}}$ exceeds $\mathrm{Avg\-TPR_{sct}}$, meaning that these models rely predominantly on dominant frequencies (corresponding to simple features) for classification of shortcut-affected classes, preventing from learning class-related semantics features. For instance, at $t{=}0.9$,  only one class (`window screen') is found subject to frequency shortcuts: ResNet18 achieves TPR=0.66 on full-spectrum images of this class and TPR=0.94 on DFM-filtered images (similarly to ResNet50 and ViT-b). 
Differently, CCT exhibits lower $\mathrm{AvgTPR^{DFM}_{sct}@0.9}$  than $\mathrm{AvgTPR_{sct}@0.9}$, indicating that while frequency shortcuts contribute significantly to classification, this model also manage to leverage other semantic information: a strong frequency shortcut (e.g. when $t{>}0.7$) does not necessarily mean that it is the only information a model uses for classification. We further compare CCT with ResNet50 as they perform similarly on IN-1k (80.57\% and 80.1\% respectively). In IN-v2 and IN-C, CCT achieves 74.81\% and 57.73\% prediction rates, having better performance than ResNet50 (74.17\% in IN-v2 and 48.85\% in IN-C).
Considering the highest degree of shortcut learning of CCT among the models (indicated by the marker size), inducing frequency reliance without blocking the learning of other semantic features appears to benefit model robustness and generalization performance under statistical distribution shifts and corruption scenarios. 

\vspace{-1em}
\paragraph{Results on IN-S and IN-R.}
Different from the results on IN-v2 and IN-C, frequency shortcuts impair generalization of models to texture and rendition changes. Model performance on shortcut and non-shortcut classes on IN-S is close to each other.  In~\cref{fig:inR_threshold}, we observe that shortcut classes have worse prediction results than non-shortcut classes when tested on IN-R which contains  images with rendition changes.   This is attributable to the fact that shortcut information are not available in the OOD tests, as the rendition and sketch test sets preserve less or very different texture information than IN-C and IN-v2 (see green lines of  IN-S and IN-1k , IN-R and IN-200 in \cref{fig:in1k_threshold} and \cref{fig:inR_threshold}). Frequency shortcut learning is an explicit cause of the texture-bias of ImageNet models, and the resulting impaired generalizability to renditions is in line with~\cite{geirhos2018imagenettrained}. 
\vspace{-1em}

\paragraph{Results on FGSM attacks.}
As shown in~\cref{fig:in1k_threshold,},  the four models, under adversarial attacks, achieve higher $\mathrm{AvgTPR}$ for shortcut classes (especially those with strong shortcuts) than non-shortcut classes. This suggests that models can be inherently  robust to adversarial noise if they leverage frequency shortcuts for classification. This is not surprising as adversarial noises hardly manipulate the textures of images.  CCT, which shows the highest degree of frequency shortcut learning among the considered models, achieves the best adversarial robustness. This indicates that model reliance on frequency shortcuts can  benefit their adversarial robustness, as the simple shortcut features are robust to adversarial noise. 
\vspace{-1em}

\begin{figure}[!t]
    % \centering
    \hspace{-1em}
    \includegraphics[width = 1.1\linewidth]{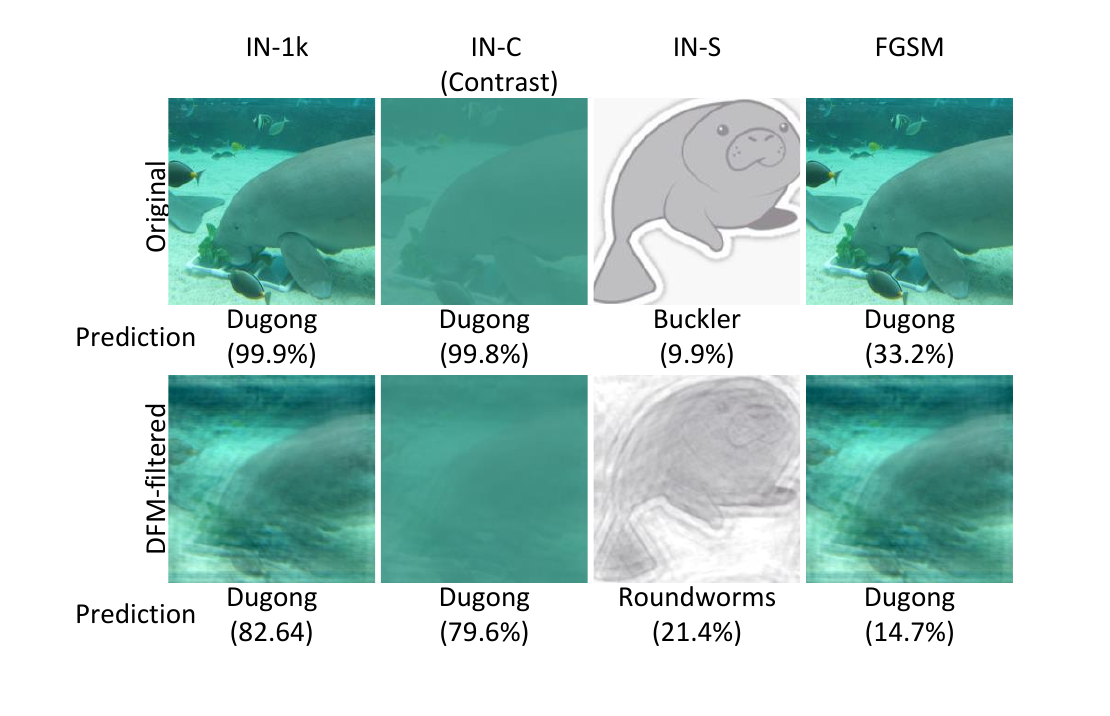}
    \vspace{-3em}
    \caption{An image of \textit{dugong} contains frequency shortcuts which a model relies on for high-confidence classification. Its presence in OOD test sets like IN-C, and in the image under FGSM attacks yields correct predictions. But the reliance on shortcuts impairs model generalization performance to renditions changes. }
    \vspace{-1em}
    \label{samples}
\end{figure}

\paragraph{Summary.}
Whether frequency shortcuts impair or benefit model generalization and robustness performance depends on the specific OOD scenarios.
% , i.e. whether textures (correspond to most frequency shortcuts) are preserved in the OOD data. 
In~\cref{samples}, we show an example image of \textit{dugong} and model predictions in OOD tests. The preserved texture information yields correct predictions on IN-C and under FGSM attacks, but it is not helpful for model generalizability to rendition changes. 
The findings reveal a limitation in the design of current OOD benchmarks, which overlooks the impact of frequency shortcuts on generalizability of models and its relation to specific  characteristics of OOD data. Using HFSS for performance evaluation can bridge such gap.

\subsection{Comparison with the existing approach}
\paragraph{More effective at finding shortcuts.} We perform experiments on IN-10 for a direct comparison with the only existing method for frequency shortcut analysis~\cite{Wang_2023_ICCV}, which %is based on individual frequency removal and 
evaluates the contribution of single frequencies to shortcuts and is limited to analyzing models trained on small-scale datasets.
We report results in~\cref{tab:compareSI}. HFSS uncovers strong frequency shortcuts for  classes \textit{airliner}, \textit{Siamese cat} (Siam-cat), \textit{ox}, \textit{frog}, \textit{zebra}, and \textit{container ship} (Con-ship),
while~\cite{Wang_2023_ICCV} was only able to find strong frequency shortcuts for classes \textit{zebra} and \textit{Siamese cat}.   The better effectiveness of HFSS at finding shortcuts is attributable to the fact that HFSS considers the joint contributions of frequencies to classification, while in~\cite{Wang_2023_ICCV} the relevance of a single frequency was measured iteratively. 
 
Frequency shortcuts can impair the generalization performance of models or  provide a false impression of good generalization when shortcuts exist in OOD tests~\cite{Wang_2023_ICCV}. We thus validate the impact of shortcuts searched by HFSS in generalization tests, by measuring performance results on IN-SCT, an OOD test set designed by~\cite{Wang_2023_ICCV}. We report the results in~\cref{tab:oodcompareSI}. The strong shortcuts for classes \textit{airliner}, \textit{frog} and \textit{container ship}   searched by HFSS  are present in the OOD set, contributing to close-to or above average TPR of classes \textit{Military aircraft} (Mil-aircraft), \textit{tree frog}, and \textit{fishing vessel} in the IN-SCT dataset.
The shortcuts searched by~\cite{Wang_2023_ICCV} for class \textit{Siamese cat} result in a much lower TPR for class \textit{tabby cat} in IN-SCT, though shortcuts for this class still exist in the OOD data (as observed from the results of HFSS). This indicates that the method proposed in~\cite{Wang_2023_ICCV} has limitations in estimating the impact of shortcuts on OOD performance, and further demonstrates the effectiveness of HFSS in finding shortcuts.

\begin{table}[!t]
\tiny  
  \caption{$\mathrm{TPR}$ results on DFM-filtered IN-10 images. $\mathrm{TPR}{\geq}0.6$ (a strong frequency shortcut) is highlighted in bold.}
  \label{tab:compareSI}
  \centering
  \renewcommand{\arraystretch}{0.85}
  \begin{tabular}{p{1.1cm}@{\hspace{1\tabcolsep}}p{2.5em} @{\hspace{1\tabcolsep}}p{2.5em} @{\hspace{1\tabcolsep}}p{2.5em} @{\hspace{1\tabcolsep}}p{2.5em} @{\hspace{1\tabcolsep}}p{3.em} @{\hspace{1\tabcolsep}}p{3.em} @{\hspace{1\tabcolsep}}p{2.5em} @{\hspace{1\tabcolsep}}p{2.5em} @{\hspace{1\tabcolsep}}p{2.5em} @{\hspace{1\tabcolsep}}p{2.5em}@{\hspace{1\tabcolsep}}}
 
     \toprule %\ \newline Method
     \bfseries   \diagbox[innerwidth=0.9cm,height=0.5cm]{Method}{Class}     & \bfseries airliner     & \bfseries wagon  & \bfseries hum-bird & \bfseries Siam-cat  & \bfseries ox & \bfseries golden \newline retriever & \bfseries frog & \bfseries zebra & \bfseries Con-ship & \bfseries  truck \\
    \midrule
    HFSS    & \textbf{0.88	}	&0	&0.52	&\textbf{0.98}	&\textbf{0.9}	&0.26	&\textbf{0.78	}&\textbf{0.7	}&\textbf{0.76}	&0.5\\
     \cite{Wang_2023_ICCV}   & 0.08 & 0 & 0.4 & \textbf{0.8} & 0.02 & 0.02 & 0.14 &\textbf{ 0.8} & 0.54 & 0.06 \\

    \bottomrule
  \end{tabular}
  \vspace{-1em}
\end{table}

\begin{table}[!t]
\tiny  
  \caption{$\mathrm{TPR}$ results on IN-SCT (first row) and corresponding DFM-filtered images (second and third rows).   $\mathrm{TPR}$ above  average TPR (0.367)  is highlighted in bold. }
  \label{tab:oodcompareSI}
  \centering
  \renewcommand{\arraystretch}{0.85}
  \begin{tabular}{p{1.1cm}@{\hspace{1\tabcolsep}}p{2.5em} @{\hspace{1\tabcolsep}}p{2.5em} @{\hspace{1\tabcolsep}}p{2.5em} @{\hspace{1\tabcolsep}}p{2.5em} @{\hspace{1\tabcolsep}}p{3.em} @{\hspace{1\tabcolsep}}p{3.em} @{\hspace{1\tabcolsep}}p{2.5em} @{\hspace{1\tabcolsep}}p{2.5em} @{\hspace{1\tabcolsep}}p{2.5em} @{\hspace{1\tabcolsep}}p{2.5em}@{\hspace{1\tabcolsep}}}
 
     \toprule
     \bfseries  \diagbox[innerwidth=0.9
     cm,height=0.5cm]{Method}{Class}   &  \bfseries Mil-aircraft     & \bfseries car  & \bfseries lorikeet & \bfseries tabby cat & \bfseries holstein & \bfseries Lab-retriever  & \bfseries  tree frog & \bfseries horse & \bfseries fishing vessel & \bfseries fire truck   \\
    \midrule
    
   --- &0.343		&\textbf{ 0.43}		&\textbf{0.443	}	&0.271		&0.329		&\textbf{0.3857}	&\textbf{0.4	}	&0.029		&\textbf{0.429	}	& \textbf{0.629	} \\
   HFSS & \textbf{0.514}	&0.029		&\textbf{0.586}	&\textbf{0.886}		&\textbf{0.457}		&0.271		&\textbf{0.871}		&0.243	&	\textbf{0.543}		&0.071 \\
    \cite{Wang_2023_ICCV}    &0      & 0      & 0.2143 & 0.1286 & 0.0429 & 0.0286 & 0.0571 & 0.1286 & 0.2143 & 0    \\
    \bottomrule
  \end{tabular}
  \vspace{-1.5em}
\end{table}
\section{Conclusions}
We proposed the first method for analysis of frequency shortcuts, HFSS, that enables the inspection of shortcut learning in large-scale models and datasets with thousands of classes. 
HFSS is more time-efficient and effective in finding frequency shortcuts compared to existing approaches. % based on individual frequency removal. 
We investigate frequency shortcut learning in models trained on ImageNet and relate their results to robustness and generalization performance under different OOD conditions. The impact of frequency shortcuts on model generalization depends on the specific OOD scenarios. 
Existing models yield good performance on generalization benchmarks when texture information, corresponding to most frequency shortcuts, is mostly preserved in OOD data.
Instead, frequency shortcuts impair the generalizability of models to images with rendition changes. This highlights the limitation of current OOD performance evaluation benchmarks, which need to explicitly take into account the impact that frequency shortcuts have on model performance. HFSS provides a tool to bridge this gap and extends the rigor of model generalization evaluation.
% \newpage
% \input{sec/ack}
% \newpage
% \section*{Acknowledgment}
% \noindent Shunxin Wang and Nicola Strisciuglio are partly supported by the ERJU project. Europe’s Rail Joint Undertaking is a European partnership on rail research and innovation established under the Horizon Europe program (2020-2027).

\small
    \bibliographystyle{ieeenat_fullname}
    \bibliography{main}
\newpage
\maketitlesupplementary
\appendix
\section{HFSS configurations}
This section details the HFSS configurations on C-10, designed to analyze the trade-off between efficiency and effectiveness in identifying shortcuts. We discuss the rationale behind the choice of frequency patch size and the percentage of sampling frequencies. 
 We then present tracking statistics of the lowest loss for an ImageNet model as frequency subsets are incrementally sampled for shortcut evaluation, which guides the selection of  $B_s$ for the HFSS configuration on IN-1k. 
\subsection{Efficiency and effectiveness of HFSS}

We perform 10 experiments on C-10, each with different number of $B_s$ as shown in~\cref{tab:experiment2}. The initial configuration is noted as CF-1 which generates in total 15000 candidate frequency subsets for shortcut relevance evaluation. The fastest (last) configuration is noted as CF-2.10, which only generates in total 70 candidate frequency subsets for evaluation. From the  results in the paper,  with CF-2.10 HFSS manages to uncover most shortcuts found by CF-1 at low thresholds. There exists a trade-off between the efficiency and effectiveness of HFSS in finding shortcuts. 

\begin{table}[h] 
\centering
\captionof{table}{Experiment configurations on C-10.}
\label{tab:experiment2}
% \tiny
  \begin{tabular}{cccccc}
    \toprule
     & \multicolumn{4}{c}{ No. of sampled candidates} &  Total  \\ 
     \cmidrule{2-5}
     CF-& $B_1$ & $B_2$ & $B_3$ & $B_4$  &  \\ 
     \midrule
     1 & 1000 & 2000 & 4000 & 8000& 15000  \\
     2.1 & 200 & 800 & 4000 & 4000& 9000  \\
     2.2 & 200 & 800 & 2000 & 2000& 5000   \\
     2.3 & 200 & 800 & 500 & 500& 2000  \\
     2.4 & 200 & 400 & 500 & 500& 1600   \\
     2.5 & 200 & 200 & 500 & 500& 1400   \\
     2.6 & 200 & 200 & 300 & 300& 1000   \\
     2.7 & 100 & 100 & 200 & 200& 600   \\
     2.8 & 50 & 50 & 100 & 100& 300 \\
     2.9 & 20 & 20 & 50 & 50& 140 \\
     2.10 & 10 & 10 & 25 & 25 & 70 \\
    \bottomrule
  \end{tabular}
\end{table}

\subsection{Patch size selection across stages}
\begin{figure}
    \centering
    \subfloat[Stage 1]{\includegraphics[width = 0.95\linewidth]{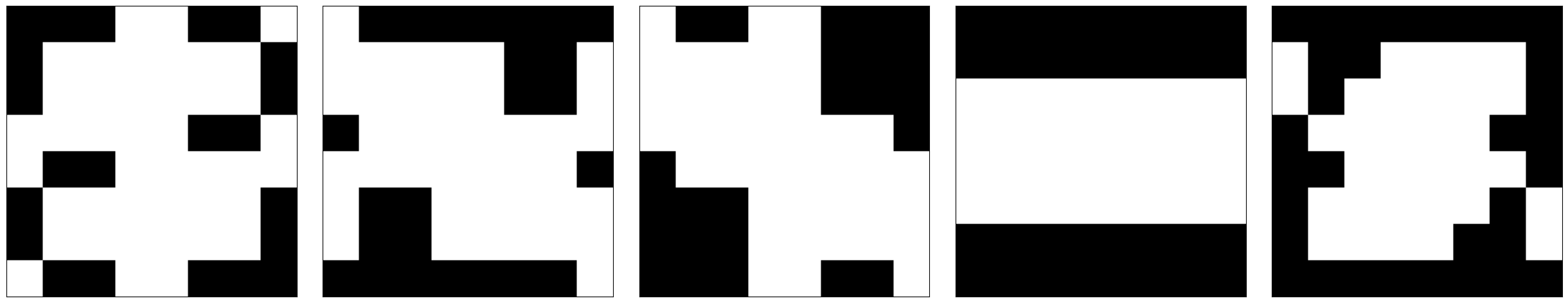}}
    % \hspace{1em}
    
    \subfloat[Stage 2]{\includegraphics[width = 0.95\linewidth]{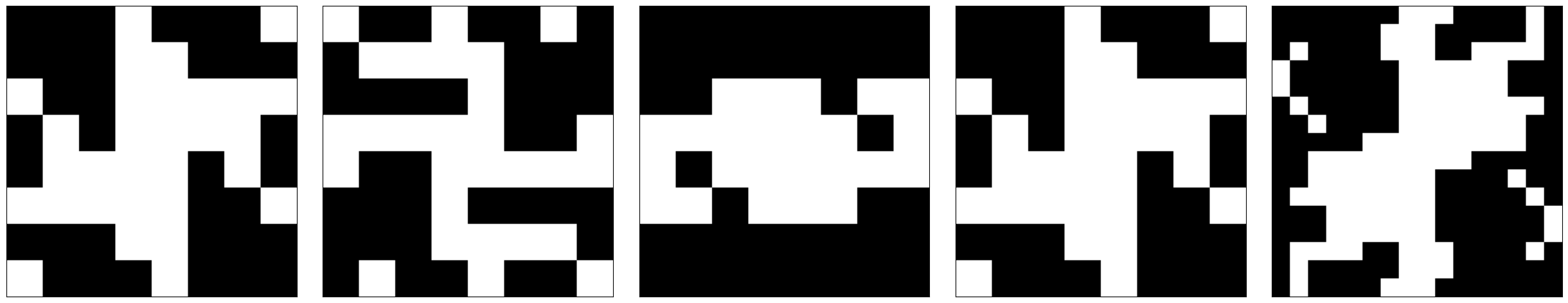}}
    
    \subfloat[Stage 3]{\includegraphics[width = 0.95\linewidth]{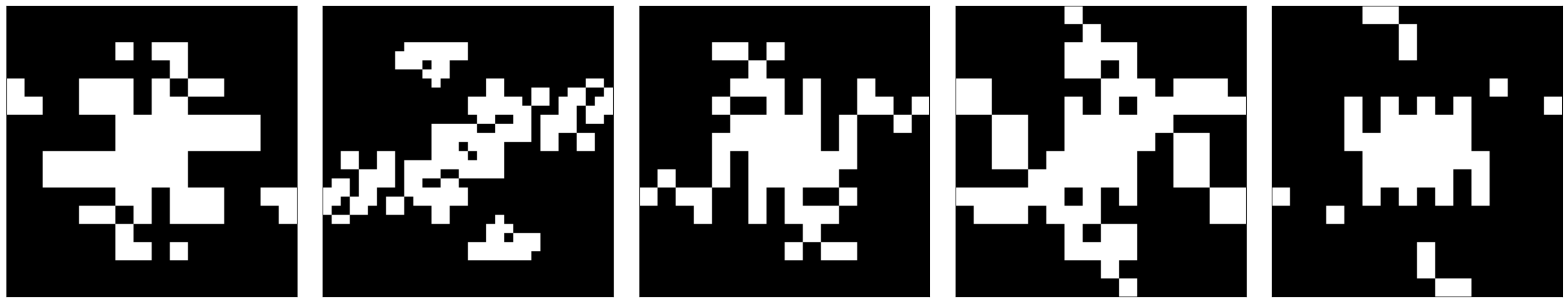}}
    % \hspace{1em}
    
    \subfloat[Stage 4]{\includegraphics[width = 0.95\linewidth]{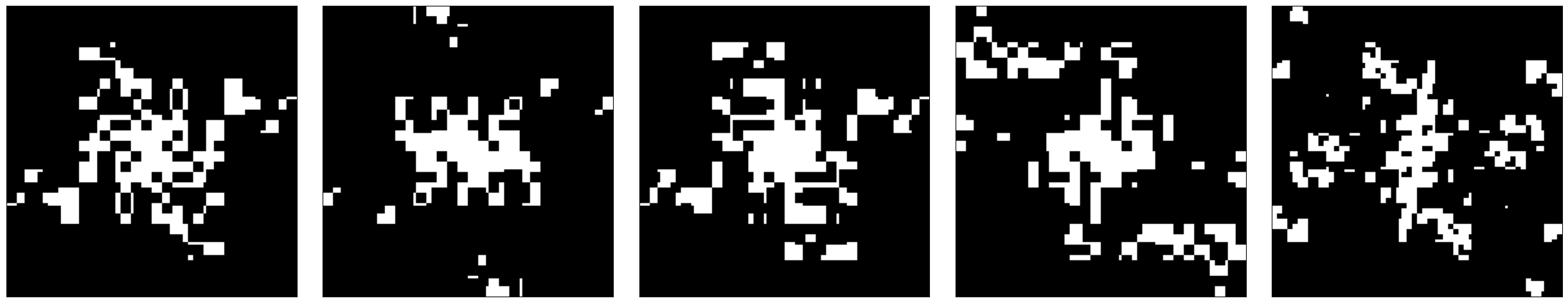}}  
    
    \subfloat[Stage 5]{\includegraphics[width = 0.95\linewidth]{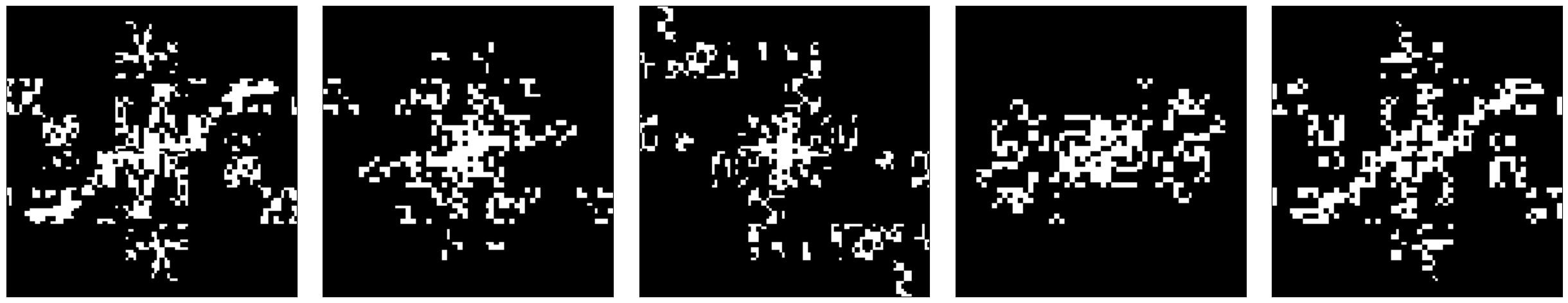}}
    % \hspace{1em}
    
    \subfloat[Stage 6]{\includegraphics[width = 0.95\linewidth]{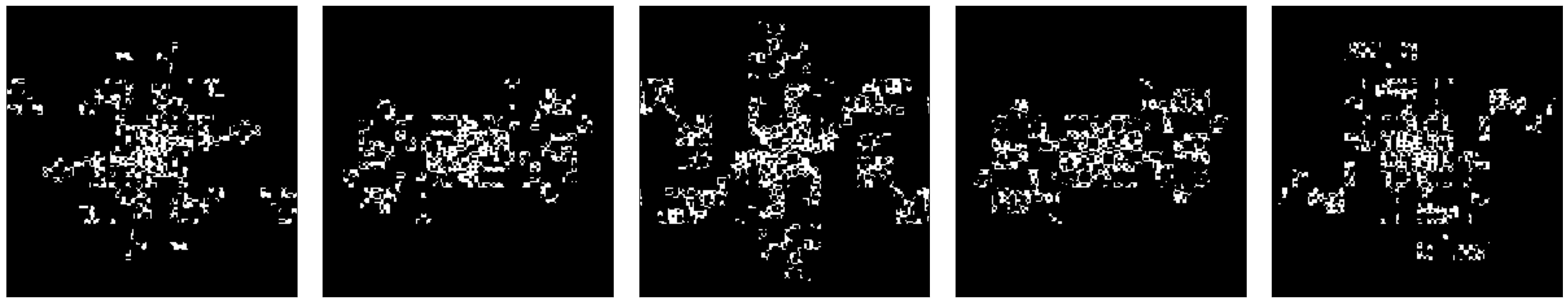}}
    \caption{Sampled frequency subsets at each stage.}
    \label{fig:samplemasks}
\end{figure}

In the first stage, we design the patch size to ensure the image spectrum can be evenly separated in $4{\times}4$ patches. This split results in a manageable number of combinations of frequency subsets, facilitating an effective initial coarse exploration of frequencies that contribute significantly to classification. From the second stage onward, the patch size is halved compared to that of the previous stage. This progressive refinement improves the precision of the frequency subsets explored that contain shortcut information. Examples of sampled frequency subsets are shown in~\cref{fig:samplemasks}. The frequency patches progressively decrease in size and the frequency maps become more refined as the search progresses.

\subsection{Frequency sampling percentage }
At each stage, we sample 60\% of the frequency patches. This uniform sampling ratio allows for investigating frequency shortcuts formed by different percentage of frequencies. For instance, applying DFMs searched at stage 3 allows us to analyze the impact of potential shortcuts that contain approximately 22\% of the frequencies on OOD data. In subsequent stages, the resulting DFMs indicate around 13\% of the frequencies across full spectrum in stage 4, and about 8\% in stage 5. 

\begin{figure}
\centering
    \subfloat[]{\includegraphics[width = 0.5\linewidth]{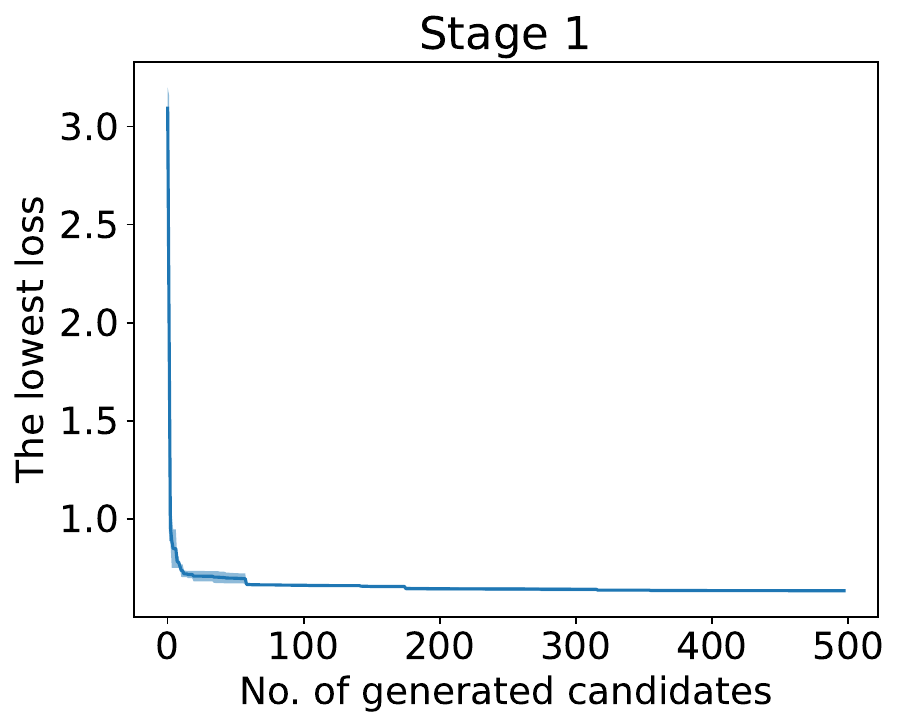}}
    \subfloat[]{\includegraphics[width = 0.5\linewidth]{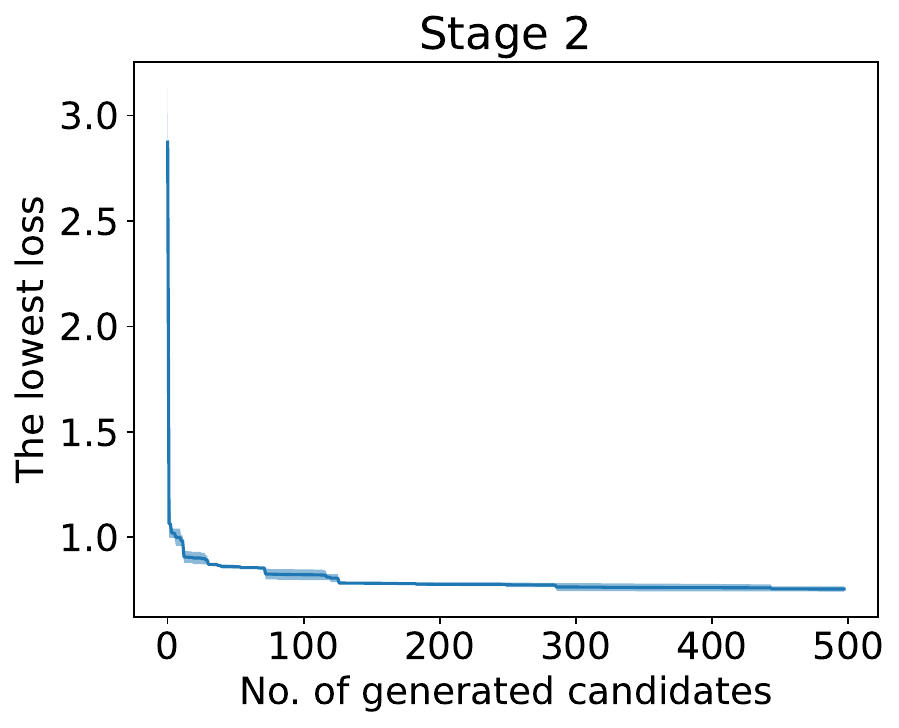}}
    
    \subfloat[]{\includegraphics[width = 0.5\linewidth]{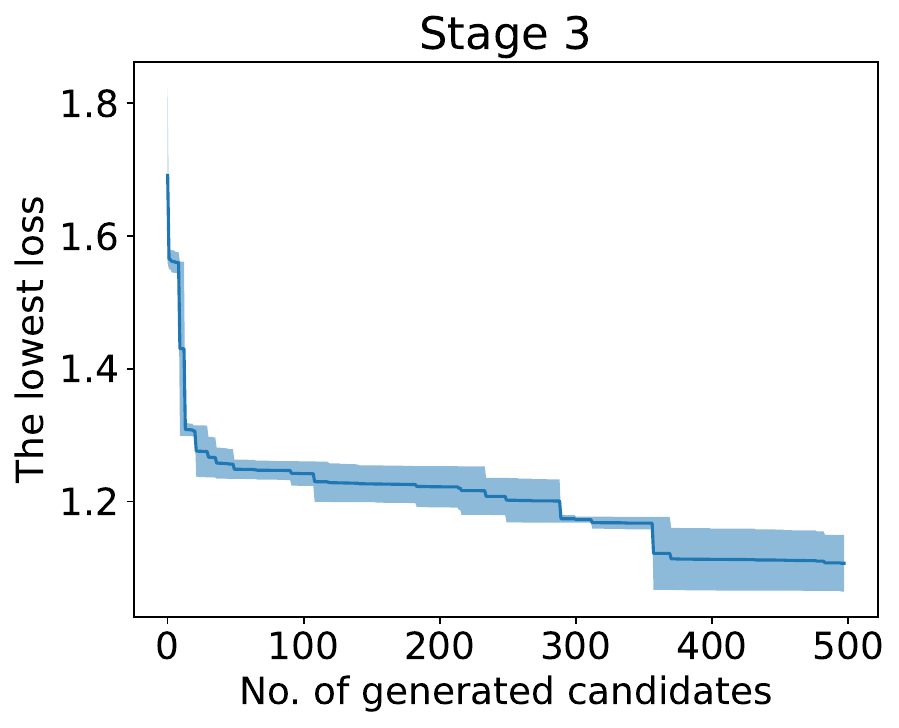}}
    \subfloat[]{\includegraphics[width = 0.5\linewidth]{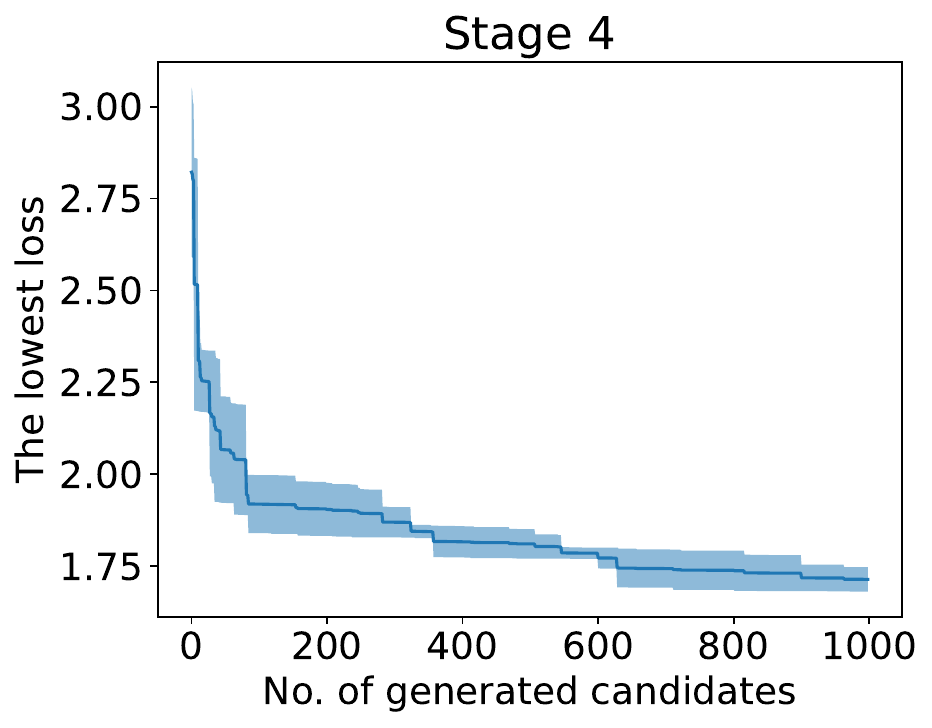}}    
    
    \subfloat[]{\includegraphics[width = 0.5\linewidth]{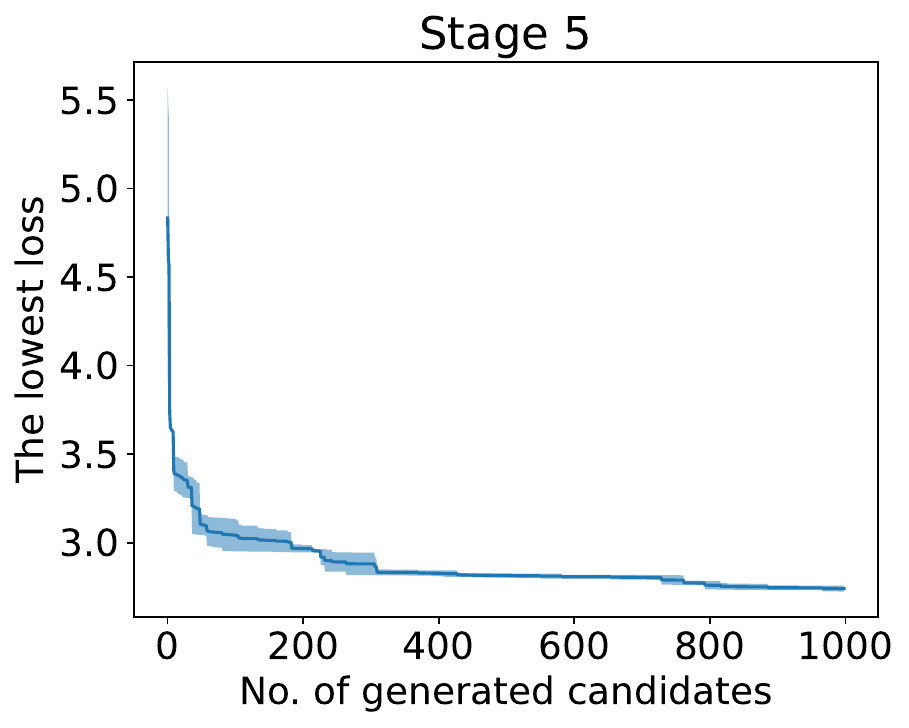}}
    \subfloat[]{\includegraphics[width = 0.5\linewidth]{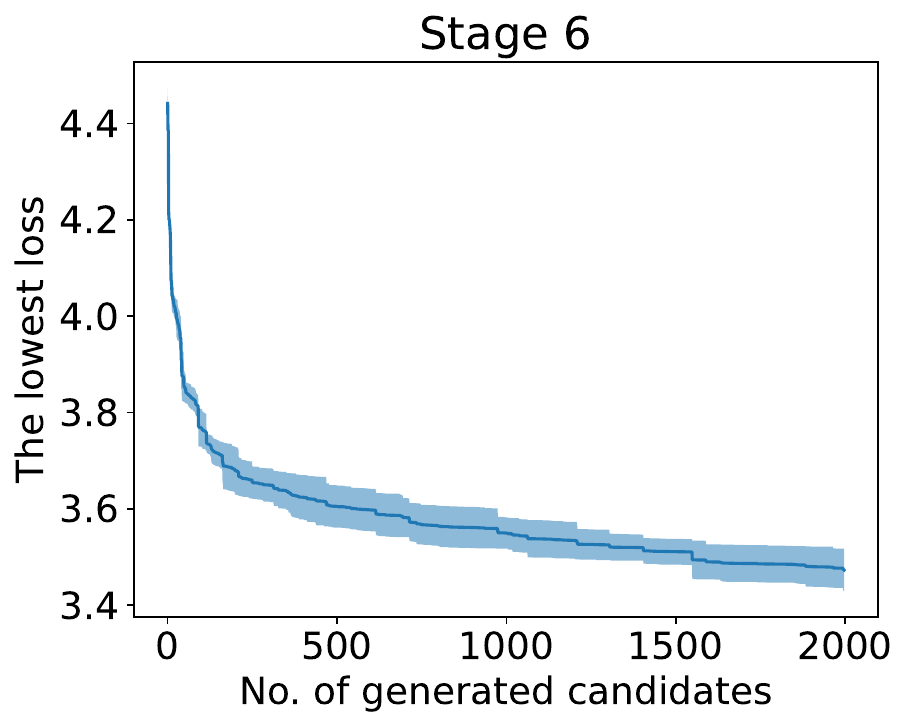}}
    \caption{Tracking statistics of IN-1k configuration of ResNet18.}
    \label{fig:track1k}
\end{figure}

\subsection{Configurations on IN-1k}
As IN-1k has 1000 classes, we increase the number of sampled candidates in each stage.
We set $B_1{=}B_2{=}B_3{=}500$, $B_4{=}B_5{=}1000$ and $B_6{=}2000$. We run this configuration three times with random seeds 42, 125 and 666. We track the  lowest loss (averaged over all 1000 classes), calculating the mean and standard deviation over the three trials. We show the tracking statistics of ResNet18 in~\cref{fig:track1k}. For stages 1 and 2, sampling 500 frequency subsets are sufficient as the loss does not decrease significantly as the number of sampled candidates increases. Starting from stage 3, although the standard deviations are relatively higher than the previous stages, the lowest loss does not decrease much after sampling more frequency subsets. Considering time-efficiency (around 9 days to run HFSS once), we use this setup for all ImageNet experiments, with slightly reduced effectiveness in finding strong shortcuts.

\section{Experiment setups}
\subsection{Datasets}

\paragraph{ImageNet-v2 (IN-v2)~\cite{pmlr-v97-recht19a}.}
This dataset has the same structure as ImageNet-1k, containing 1000 classes. The data creation process of IN-v2 is the same as that of IN-1k. This can evaluate model performance on images collected in different time points, i.e. generalization to statistical distribution shifts. 

\paragraph{ImageNet-C (IN-C)~\cite{hendrycks2018benchmarking}.}  It contains 19 types of synthetic corruption effect, which are Gaussian noise, impulse noise, shot noise, defocus blur, glass blur, motion blur, zoom blur, brightness, contrast, elastic transform, jpeg compression, pixelate, fog, frost, and snow. The dataset contains 19 subsets, each containing IN-1k test images corrupted by one type of corruption, with five levels of corruption severity. High severity indicates high strength of corruption applied to the original test images.

\paragraph{ImageNet-R (IN-R)~\cite{Hendrycks_2021_ICCV}.} 
It contains images with different renditions, such as  cartoon, art, graphics, painting, etc. These allow for a strong  model generalizability assessment, as some abstract renderings exclude important features like natural textures that models rely on for classification.

\paragraph{ImageNet-S (IN-S)~\cite{wang2019learning}.}
The dataset contains 1000 classes, each with 50 validation images, the same as IN-1k. Differently, the images are sketches of objects, which may have texture information loss. 

\paragraph{ImageNet-SCT (IN-SCT)~\cite{Wang_2023_ICCV}.}
This OOD dataset is constructed to evaluate the impact of frequency shortcuts on generalization performance. It contains 10 classes, sharing similar shape or texture characteristics to the 10 classes in IN-10. Each class has 70 images with seven renditions, e.g. cartoon, painting, sketch, etc.

\subsection{Training}
\paragraph{C-10.}
Models with ResNet~\cite{He_2016_CVPR} architecture are trained for 200 epochs on the C-10 dataset. The initial learning rate is 0.01, reduced by a factor of 10 if the validation loss does not decrease for 10 epochs. We use SGD optimizer with momentum 0.9 and weight decay $10^{-4}$ and batch size 128.

\paragraph{IN-10.}
Models with ResNet(s)~\cite{He_2016_CVPR}  architectures are trained for 200 epochs. The initial learning rate is 0.01 and is reduced by a factor of 10 if the validation loss does not decrease for 10 epochs. We use SGD optimizer with momentum 0.9 and weight decay $10^{-4}$ and batch size 16. 

\begin{figure*}[!t]
    \centering
    \subfloat[]{\includegraphics[width=0.85\linewidth]{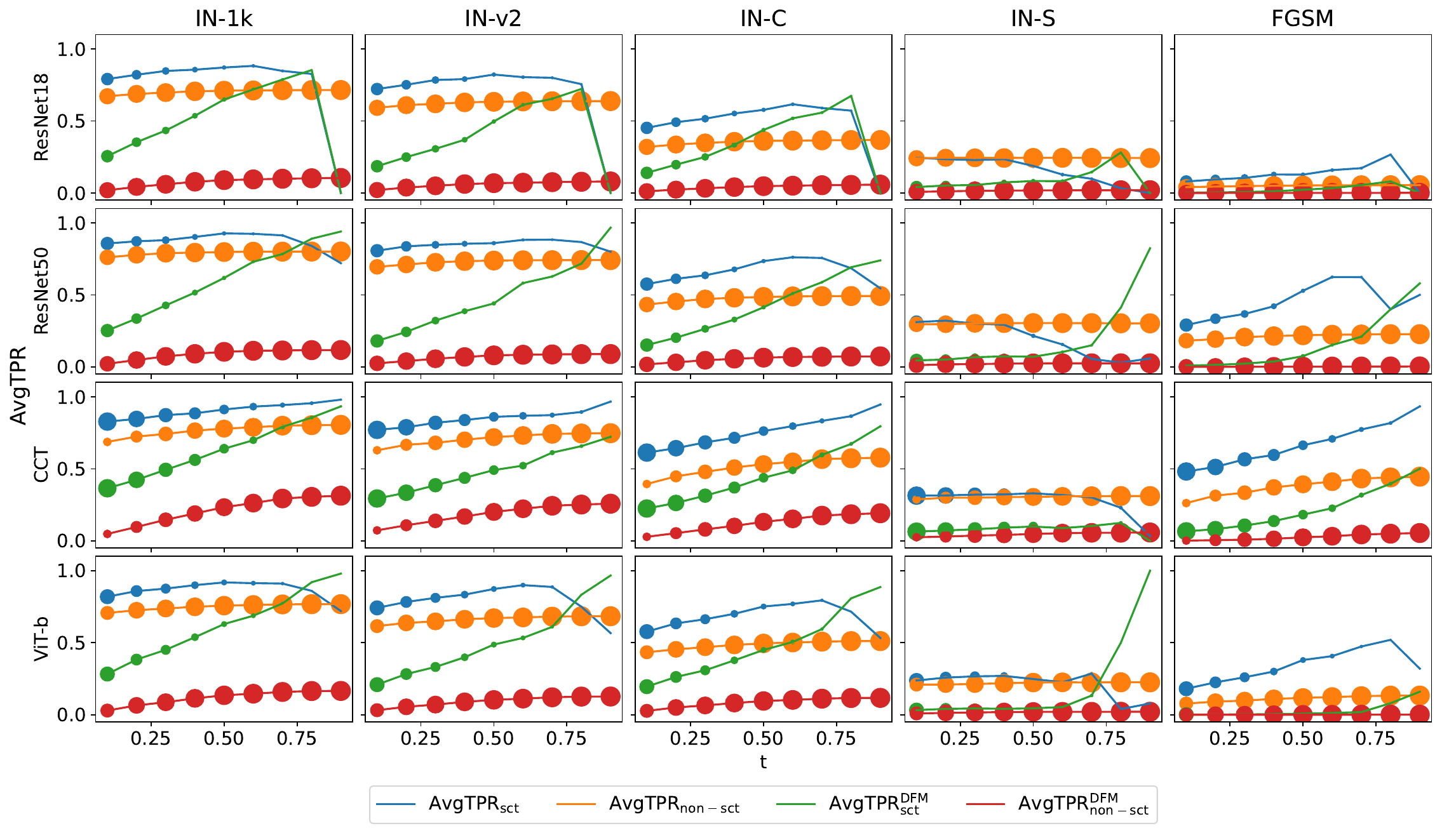} \label{fig:T2_s6}}

    \subfloat[]{\includegraphics[width=0.85\linewidth]{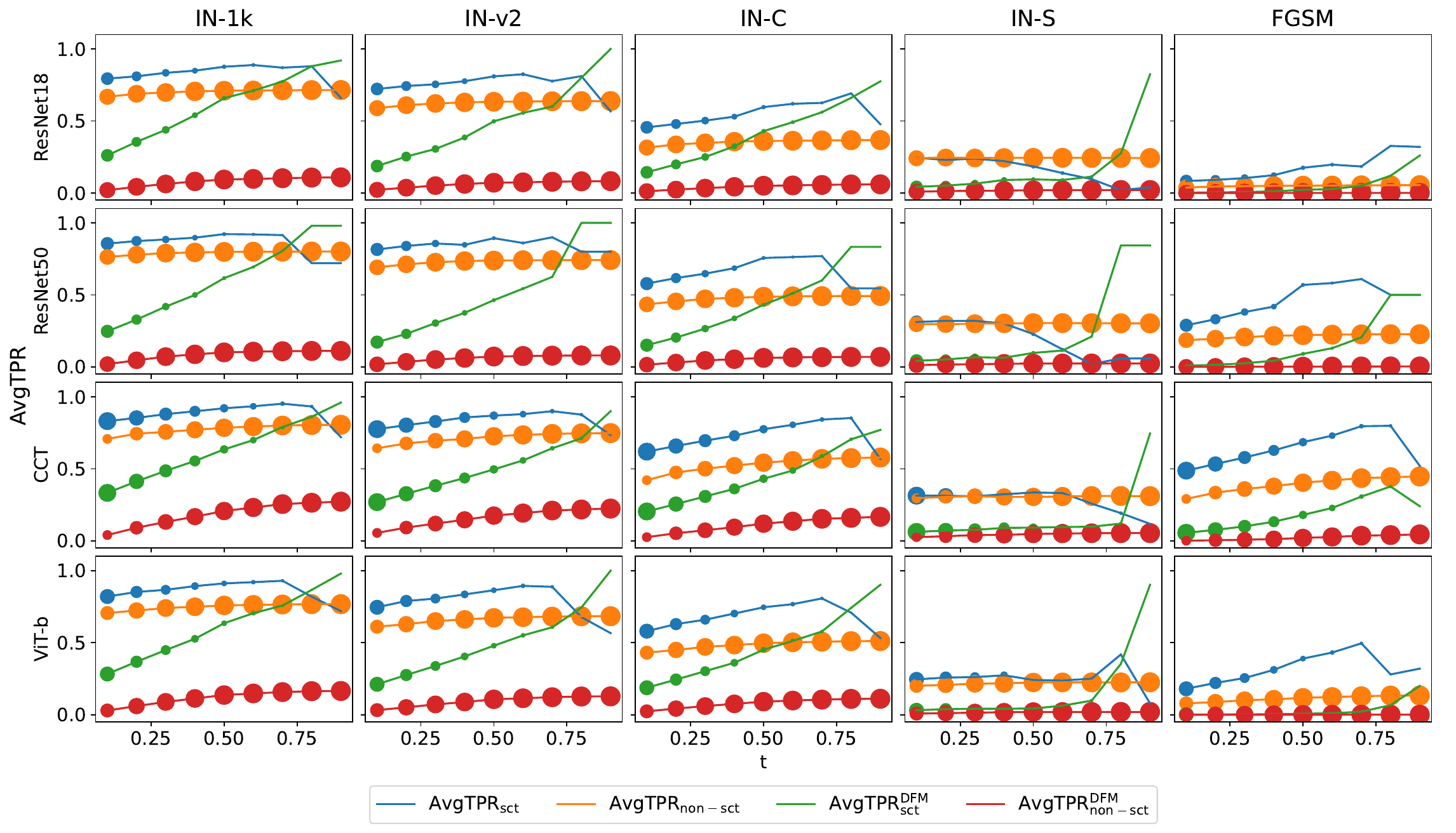}     \label{fig:T3_s6}}
    \caption{Impact of shortcuts uncovered in (a) the second and (b) the third run on OOD data: average TPR of shortcut and non-shortcut classes given different thresholds. In general, models perform better on images of shortcut classes than non-shortcut classes.   }
    \label{fig:T2_T3}
\end{figure*}

\paragraph{IN-1k.} We use the pre-trained weights of ResNet18, ResNet50 and ViT-b from timm~\cite{Wightman} and the weights of CCT from the official repository~\cite{hassani2021escaping}.

\section{Additional results}

\begin{figure*}
    \centering
    \subfloat[]{\includegraphics[width=0.5\linewidth]{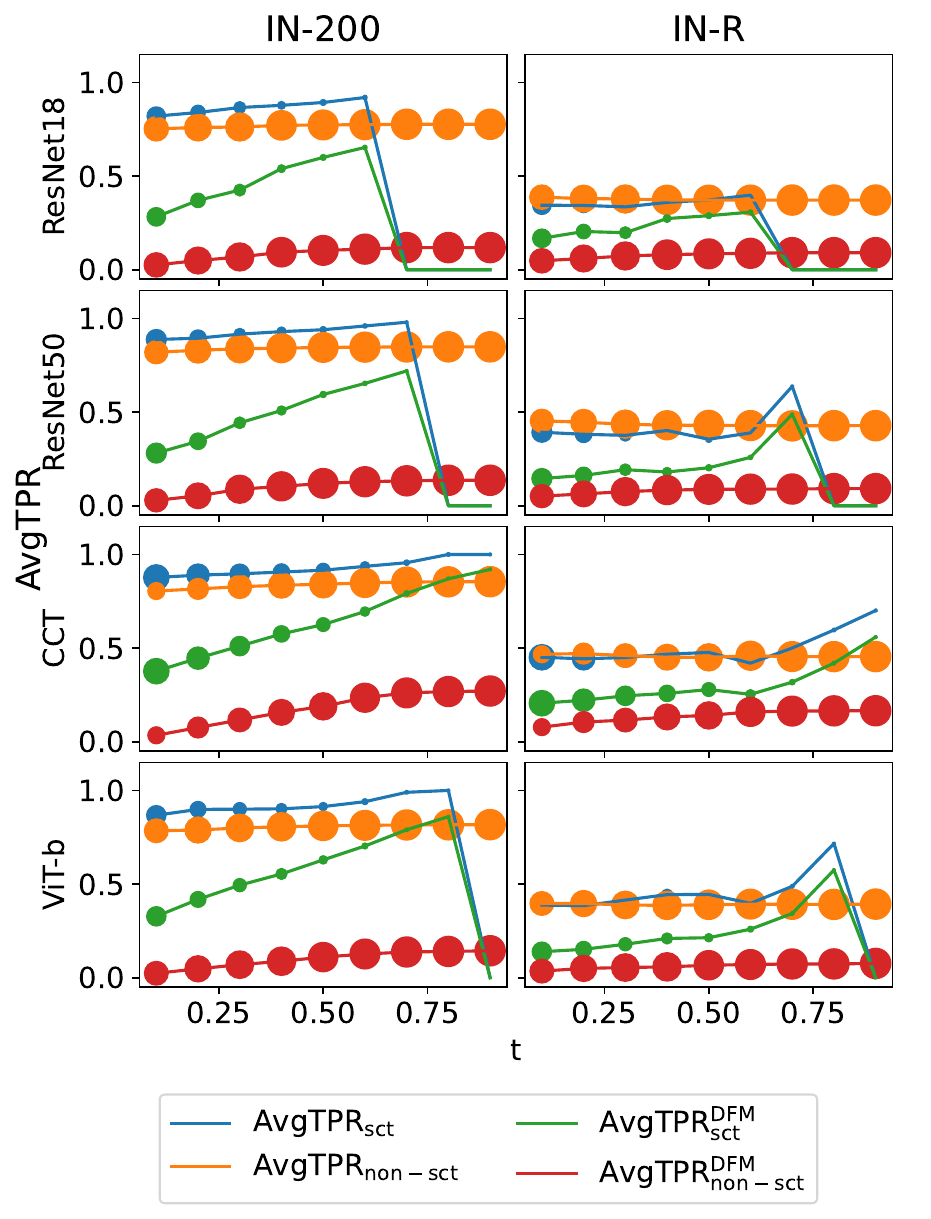}}
    \subfloat[]{\includegraphics[width=0.5\linewidth]{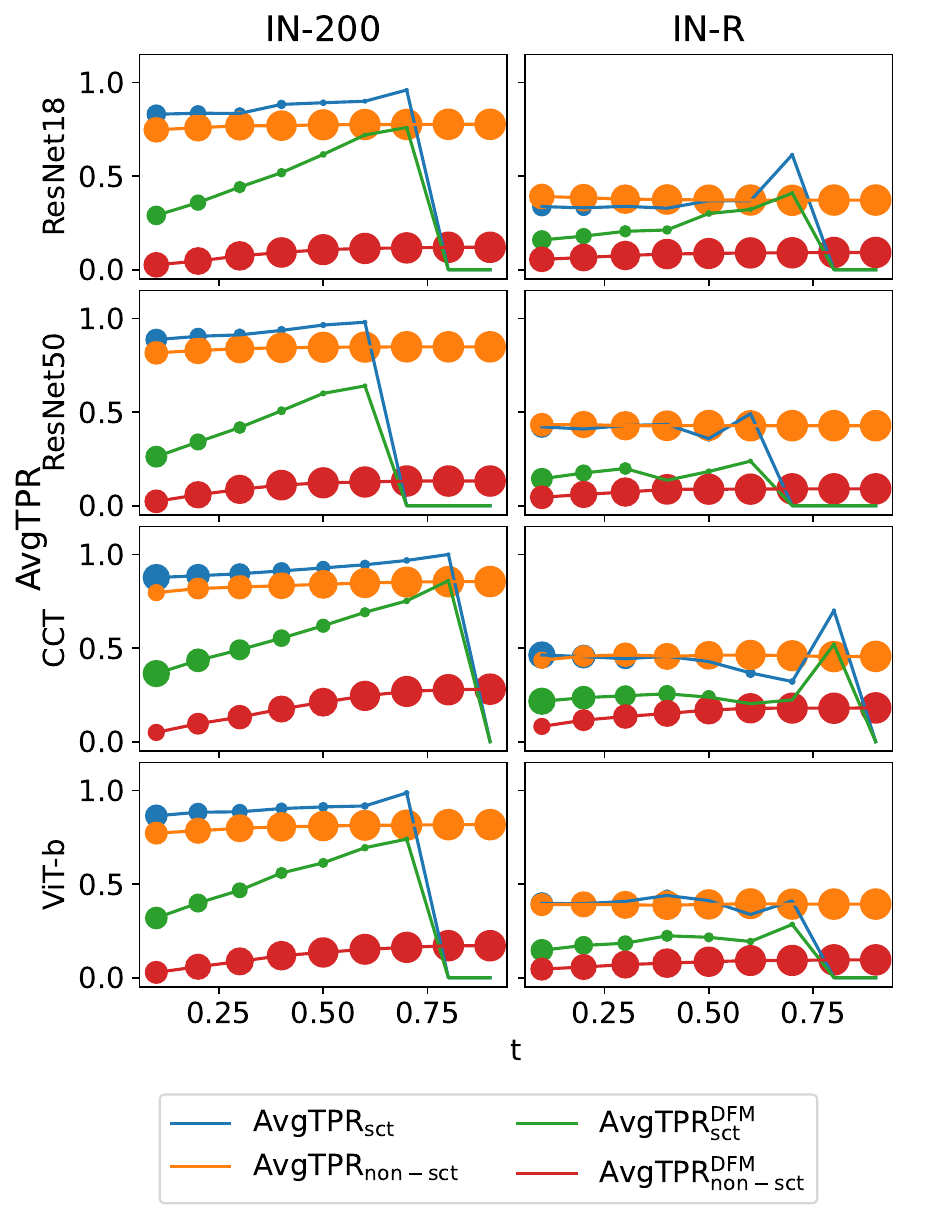}}
    \caption{Impact of shortcuts uncovered in (a) the second and (b) the third run on IN-R: average TPR of shortcut and non-shortcut classes given different thresholds. In general, models perform worse on images of shortcut classes than non-shortcut classes.}
    \label{fig:T2T3_R}
\end{figure*}

\subsection{IN-1k}
\paragraph{Results across multiple runs.}
We conducted ImageNet experiments for each model three times, with results from the additional two trials presented in~\cref{fig:T2_T3,fig:T2T3_R}. 

Similar to the findings from the first run, the subsequent trials also manage to identify classes influenced by shortcuts. Models consistently perform better on shortcut-classes than non-shortcut classes across datasets such as IN-1k, IN-v2, IN-C and under FGSM attacks but worse on IN-R. These results align with the observations that models  excel on texture-preserved datasets,  as they exploit the shortcuts present in OOD data.   Notably, CCT shows the strongest tendency toward shortcut learning among the evaluated models.

Although the current HFSS configuration applied to ImageNet might overlook some strong shortcuts (see the green lines of ResNet18 in~\cref{fig:T2_s6}), HFSS stably uncovers shortcuts at low thresholds (weak shortcuts). As our focus is on the general impact of shortcuts on generalization and robustness, rather than precise prediction performance on specific classes, the configuration of HFSS provides analyzable results for such investigation. For more detailed analyses, one could increase the number of sampling operations $B_s$, allowing a broader evaluation of frequency subset combinations and obtaining more stable search of strong shortcuts. 

% \newpage

 \begin{figure*}
    \centering
     \subfloat[]{\includegraphics[width=0.7\linewidth]{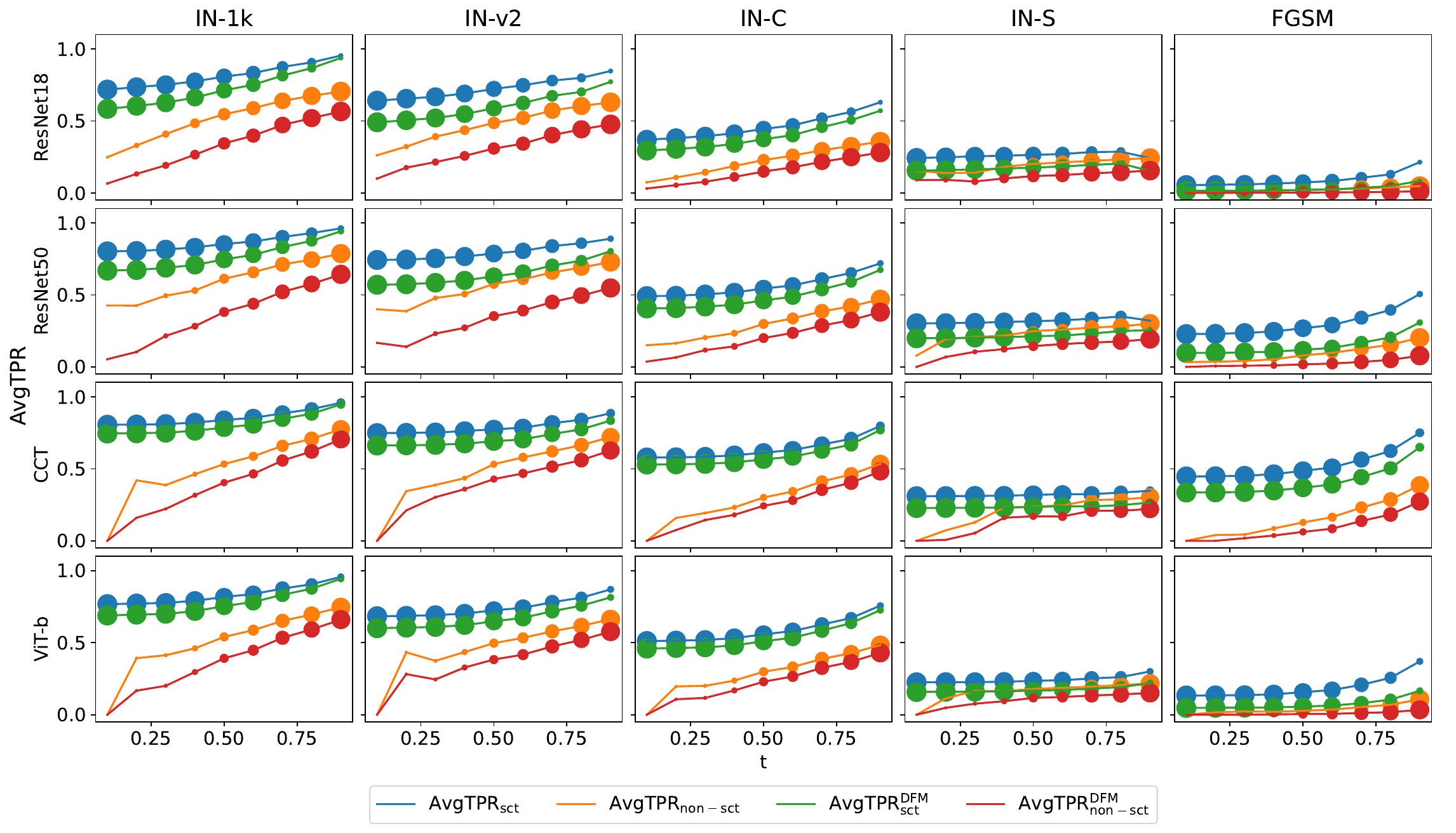}   \vspace{-0.5em} \label{fig:T1_s3}}

    \subfloat[]{\includegraphics[width=0.7\linewidth]{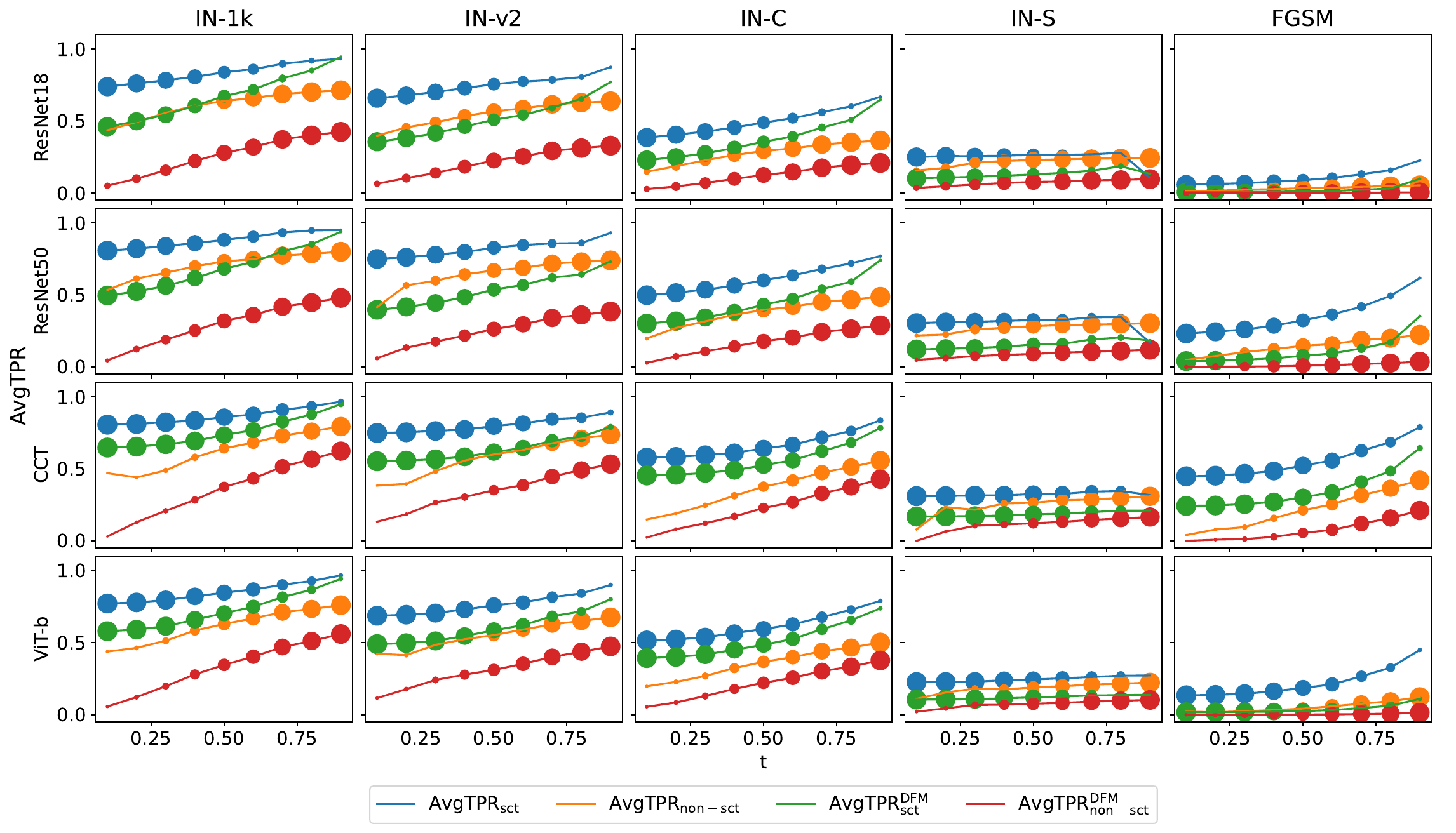}  \vspace{-0.5em} \label{fig:T1_s4}}
    
    \subfloat[]{\includegraphics[width=0.7\linewidth]{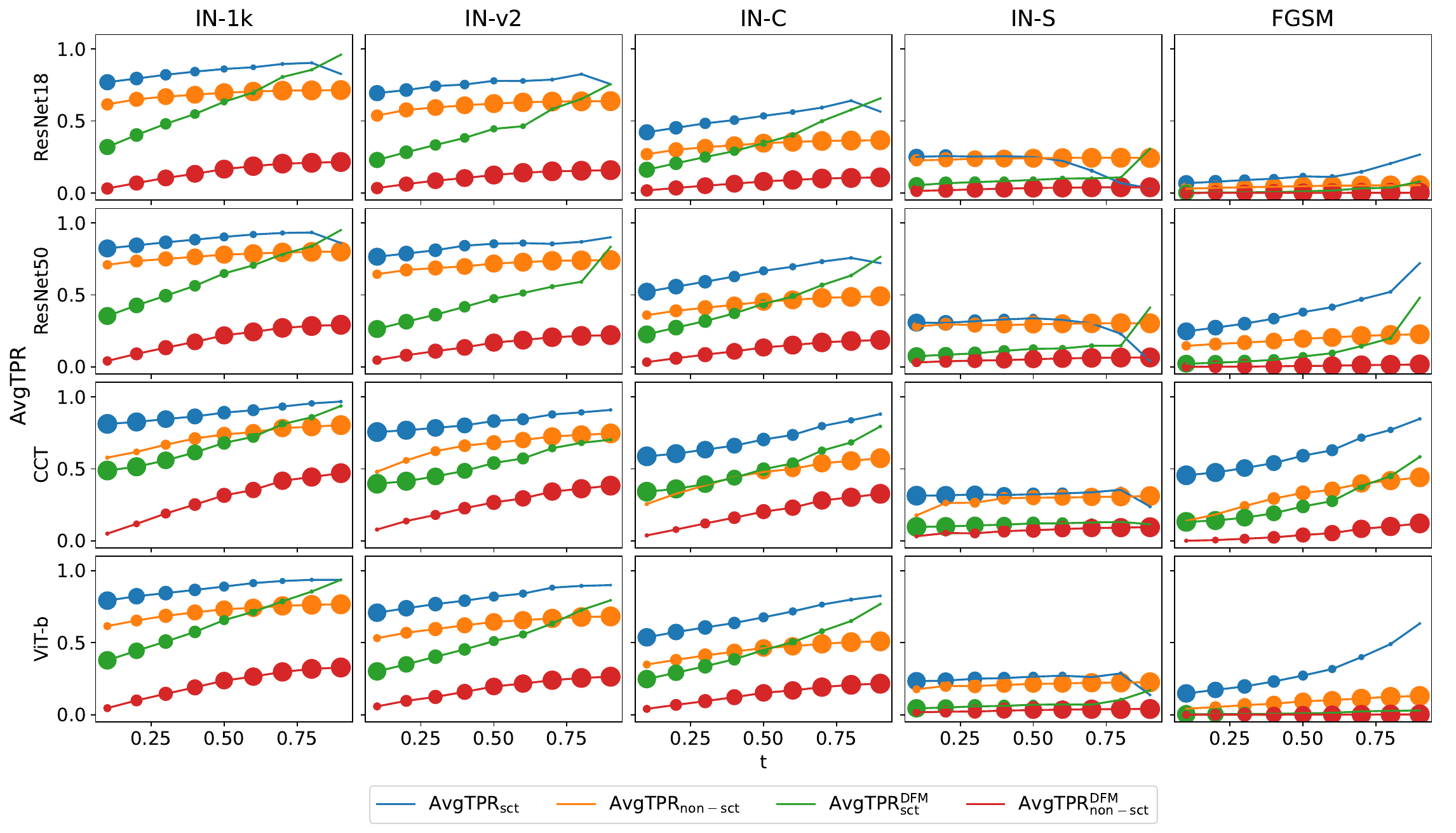}  \vspace{-0.5em} \label{fig:T1_s5}}
     \vspace{-1em}
    \caption{Average TPR of shortcut and non-shortcut classes given different thresholds, using DFMs containing around (a) 22\%, (b) 13\% and (c)  8\% of frequencies. In general, models perform better on images of shortcut classes than non-shortcut classes.}
    \label{fig:T1}
\end{figure*}

 \begin{figure*}
     \centering
     \subfloat[]{\includegraphics[width=0.33\linewidth]{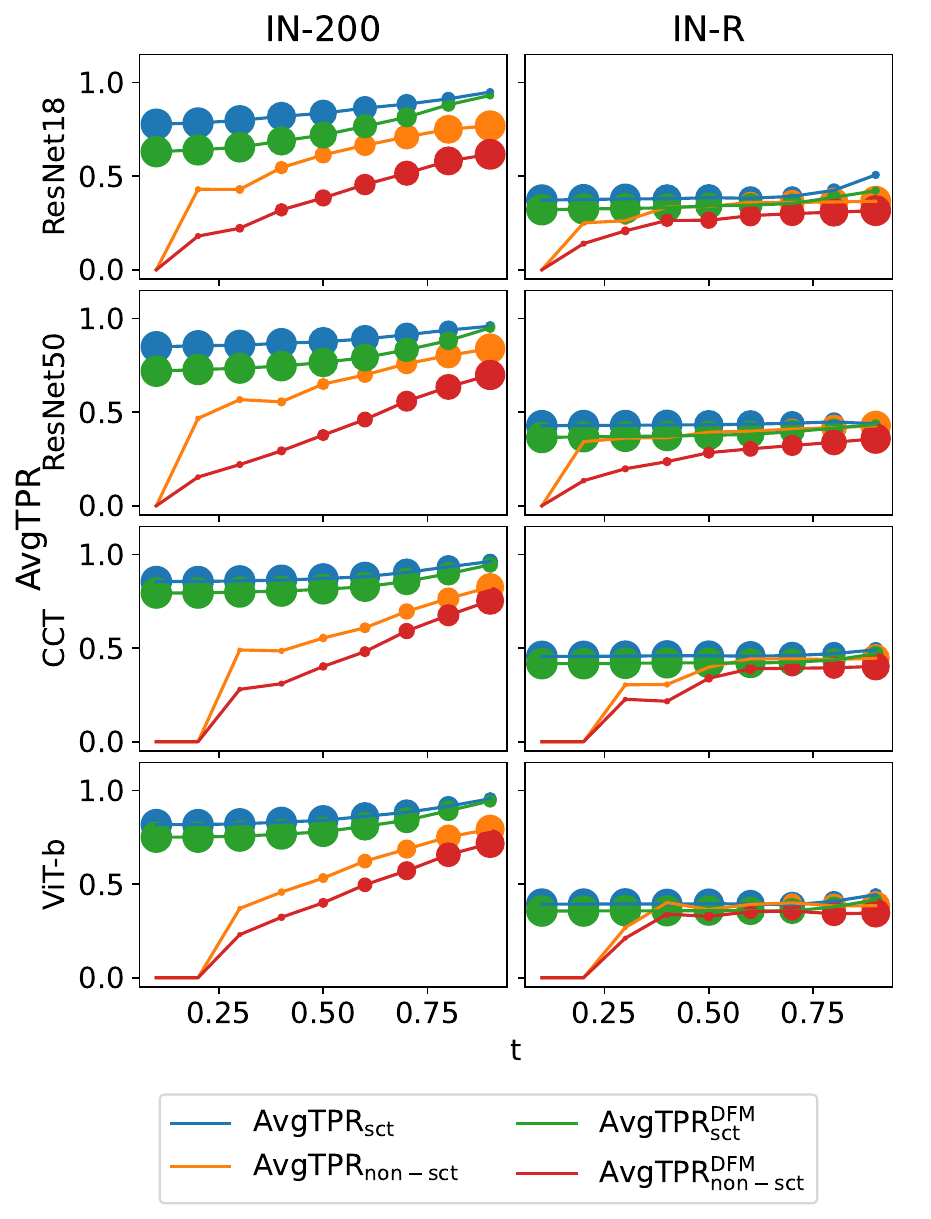} \label{fig:T1_R3}}
    \subfloat[]{\includegraphics[width=0.33\linewidth]{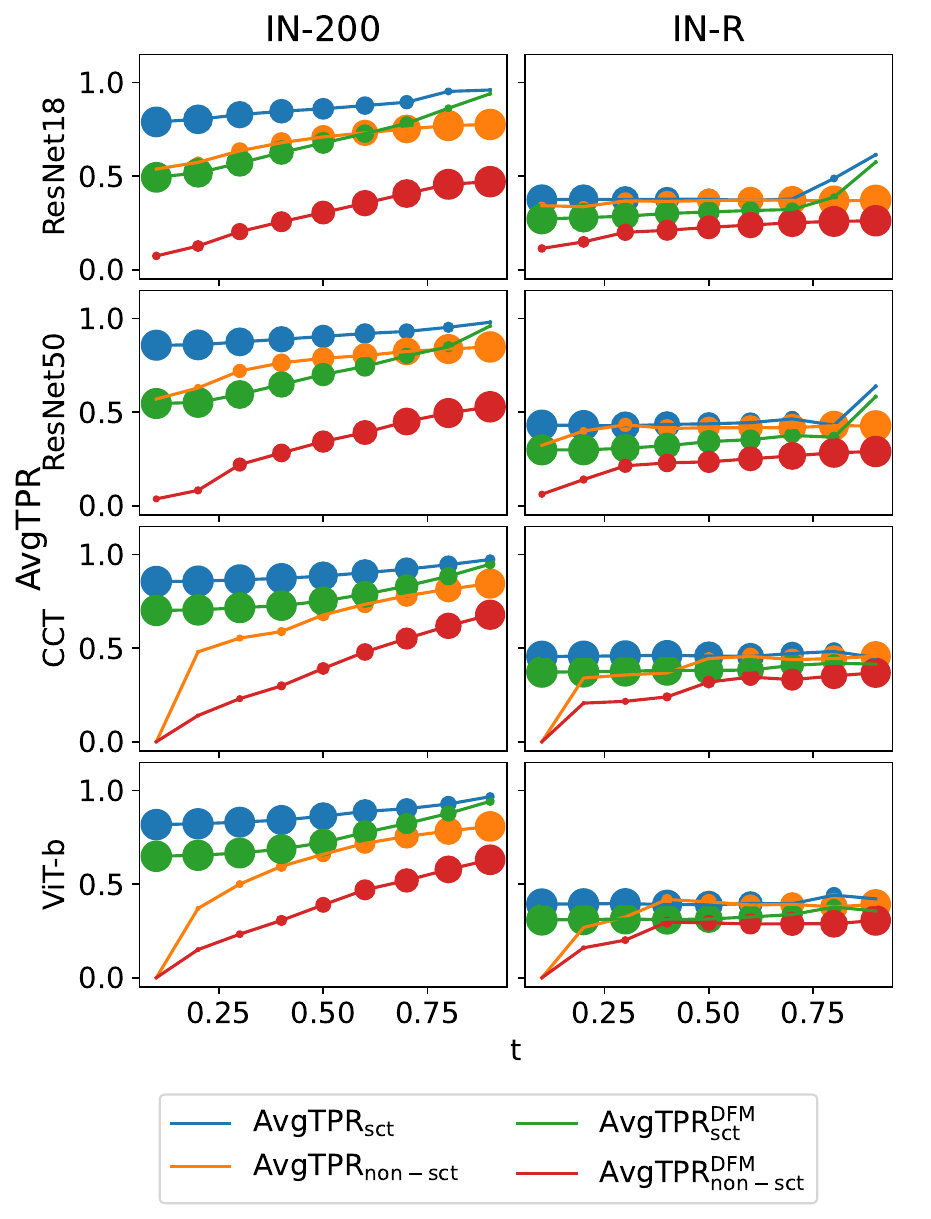}}
    \subfloat[]{\includegraphics[width=0.33\linewidth]{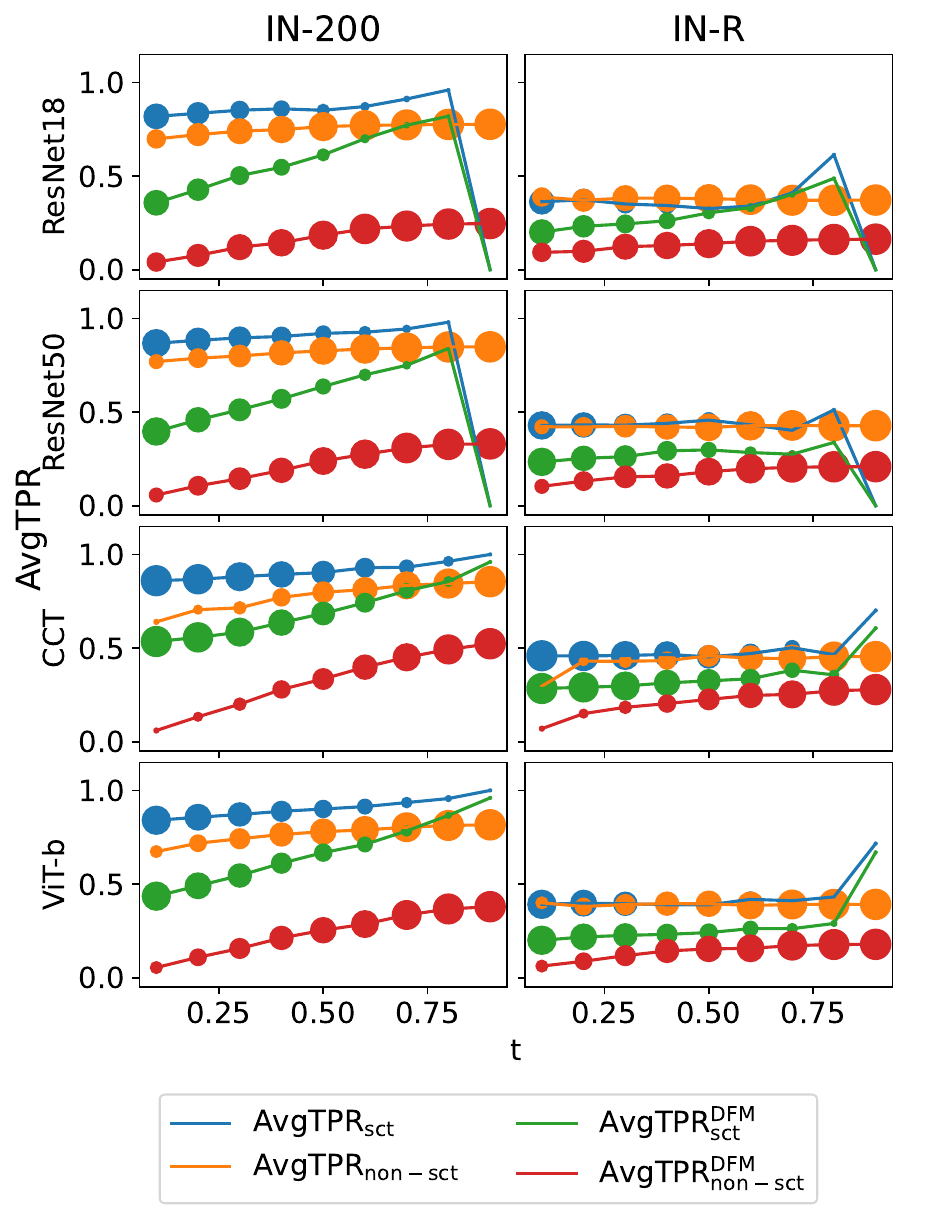}}
     \caption{Average TPR of shortcut and non-shortcut classes given different thresholds on IN-R, using DFMs containing around (a) 22\%, (b) 13\% and (c)  8\% of frequencies. In general, models perform similarly on images of shortcut classes and non-shortcut classes.}
     \label{fig:T1_R}
 \end{figure*}
% \newpage

\paragraph{Results at different stages.}
We analyze the impact of frequency shortcuts searched in different stages, which correspond to different percentage of frequencies, on the ID and OOD test sets, as shown in~\cref{fig:T1,fig:T1_R}. Sampling $60\%$ frequency patches at each stage results in around 22\% of frequencies  at stage 3,  13\% at stage 4 and 8\% at stage 5 of the full image spectrum. 

 In~\cref{fig:T1_s3,fig:T1_R3}, we present the average TPR of models using DFMs obtained at stage 3 (approximately 22\% of frequencies). For shortcut classes, the models achieve performance comparable to that on full-spectrum images. But for non-shortcut classes, performance is generally worse. On IN-S and IN-R datasets, models perform slightly better on shortcut classes than non-shortcut classes at low thresholds.  We inspect images filtered by the DFMs from stage 3 and see that retaining only 22\% of frequencies results in minimal visual differences compared to the original images, aside from some artifacts caused by filtering. As this retained information is sufficient for classification, most classes are considered subject to shortcuts at low thresholds. This explains the slightly higher Average TPR values, particularly for $\mathrm{AvgTPR}@0.1$ in CCT (see~\cref{fig:T1_s3}), where almost no non-shortcut classes remain.
 Despite this, the performance decline of shortcut classes from IN-1k to In-S and IN-R is notably more pronounced compared to non-shortcut classes, showing that reliance on frequency shortcuts does not aid model generalization.
 The larger performance drops on IN-S (compared to the drop on IN-R) can also be attributed to  shortcuts such as color-related cues, as IN-S only contains black-and-white sketches.  Similar trends are observed for stages 4 and 5, as shown in~\cref{fig:T1,fig:T1_R}. 

 \paragraph{Number of classes per stage.}
\cref{tab:noofclassIN1k} presents the number of shortcut classes at each stage. As the threshold $t$ increases, the count of shortcut classes declines. Notably, at stage 6 with $t=0.9$, the number drops to 1-3, indicating that such strong shortcuts are uncommon.

\begin{table}[!t]
\tiny 
  \caption{ The number of shortcut classes per stage in IN-1k.}
  \label{tab:noofclassIN1k}
  \centering
  \begin{tabular}{p{1.1cm}@{\hspace{1\tabcolsep}}p{2.5em} @{\hspace{1\tabcolsep}}p{2.5em} @{\hspace{1\tabcolsep}}p{2.5em} @{\hspace{1\tabcolsep}}p{2.5em} @{\hspace{1\tabcolsep}}p{3.em} @{\hspace{1\tabcolsep}}p{3.em} @{\hspace{1\tabcolsep}}p{2.5em} @{\hspace{1\tabcolsep}}p{2.5em} @{\hspace{1\tabcolsep}}p{2.5em} @{\hspace{1\tabcolsep}}}
    \toprule
  
      \bfseries   \diagbox[innerwidth=0.9cm,height=0.5cm]{Stage}{$t$} & \bfseries 0.1     & \bfseries 0.2  & \bfseries 0.3 & \bfseries  0.4 & \bfseries 0.5 & \bfseries 0.6 & \bfseries  0.7 & \bfseries 0.8 & \bfseries 0.9   \\
    \midrule
    \bfseries ResNet18 \\
    \bfseries  3     & 993 &950 & 894 &792 &642 &516 &317 & 175 & 40\\
    \bfseries  4     & 922 &826  & 701 &539 & 381& 276& 133&62  &9 \\
    \bfseries  5      & 647 &445 & 301 &203 & 111& 61&23 &12  &3 \\
    \bfseries  6      &  343& 196&116  & 60& 22&16 &7 &3  &1 \\
    \bfseries ResNet50 \\
    \bfseries 3 & 997 &990 &956 &904 &782 &673 &469 &305 &82 \\
    \bfseries 4 & 976& 903& 792& 639& 457& 339& 175& 95& 12\\
    \bfseries 5 & 809& 604& 445& 308& 174& 110& 45& 18& 2\\
    \bfseries 6 & 421& 253& 149& 78& 33& 18& 5&1 &1 \\
    \bfseries CCT \\
    \bfseries 3 & 1000& 997& 989& 958& 890& 820& 648& 475& 166\\
    \bfseries 4 & 998& 985& 948& 885& 755& 636& 421& 256& 72\\
    \bfseries 5 & 973& 906& 778& 622& 442& 336& 152& 82& 15\\
    \bfseries 6 & 781& 566& 408& 277& 142& 88& 34&15 &3 \\
    \bfseries ViT-b \\
    \bfseries 3 & 1000& 989& 977& 930&820 &720 &518 &343 &101 \\
    \bfseries 4 & 987& 962& 900& 772& 633& 491& 297& 171& 33\\
    \bfseries 5 & 861& 674& 514& 368& 225& 149& 70& 25& 6\\
    \bfseries 6 & 520 &317 &183 &110 &54 &30 &7 &1 &1 \\
    \bottomrule
  \end{tabular}
\end{table}

\begin{table}[!t]
\tiny 
  \caption{ TPRs on  C-10  and $\mathrm{DFM}$-filtered images.  $\mathrm{TPR}{\geq}0.6$  are highlighted in bold. }
  \label{tab:idin10}
  \centering
  \begin{tabular}{p{1.1cm}@{\hspace{1\tabcolsep}}p{2.5em} @{\hspace{1\tabcolsep}}p{2.5em} @{\hspace{1\tabcolsep}}p{2.5em} @{\hspace{1\tabcolsep}}p{2.5em} @{\hspace{1\tabcolsep}}p{3.em} @{\hspace{1\tabcolsep}}p{3.em} @{\hspace{1\tabcolsep}}p{2.5em} @{\hspace{1\tabcolsep}}p{2.5em} @{\hspace{1\tabcolsep}}p{2.5em} @{\hspace{1\tabcolsep}}p{2.5em}@{\hspace{1\tabcolsep}}}
    \toprule
  
      \bfseries   \diagbox[innerwidth=0.9cm,height=0.5cm]{Model}{Class} & \bfseries airplane     & \bfseries auto  & \bfseries bird & \bfseries  cat & \bfseries deer & \bfseries dog & \bfseries  frog & \bfseries horse & \bfseries ship & \bfseries truck  \\
    \midrule
    \bfseries  ResNet18     & \textbf{0.709	} &	0.581		& \textbf{0.86}&	0.585		&\textbf{0.679}		&0.482		&\textbf{0.713	}&0.395		& \textbf{0.675}	& \textbf{0.682}  \\
     
  \bfseries  ResNet34   &\textbf{ 0.988}	& 0.183		&\textbf{0.935}	&\textbf{0.641}		&0.585		&0.538		&0.393	&	0	&0.203	&0.467	 \\    
    
    \bfseries  ResNet50   & \textbf{0.995	}	&	0.365		&	\textbf{0.858	}	&	0.526		&	0.361		&	0.432		&	0.251		&	0.259		&	0.338		&	0.526	 \\

    \bottomrule
  \end{tabular}
\end{table}

\begin{table}[!t]
\tiny 
  \caption{TPRs of C-10 models tested on resized IN-10 (first row of each model) and corresponding DFM-filtered images (second row of each model). TPRs higher than or close to average TPR (ResNet18-0.62, ResNet34-0.62 and ResNet50-0.64) are highlighted in bold. }
  \label{tab:oodcifar}
  \centering
  \renewcommand{\arraystretch}{0.85}
 \begin{tabular}{p{1.1cm}@{\hspace{1\tabcolsep}}p{2.5em} @{\hspace{1\tabcolsep}}p{2.5em} @{\hspace{1\tabcolsep}}p{2.5em} @{\hspace{1\tabcolsep}}p{2.5em} @{\hspace{1\tabcolsep}}p{3.em} @{\hspace{1\tabcolsep}}p{3.em} @{\hspace{1\tabcolsep}}p{2.5em} @{\hspace{1\tabcolsep}}p{2.5em} @{\hspace{1\tabcolsep}}p{2.5em} @{\hspace{1\tabcolsep}}p{2.5em}@{\hspace{1\tabcolsep}}}
 
     \toprule %\ \newline Method
     \bfseries   \diagbox[innerwidth=0.9cm,height=0.5cm]{Method}{Class}     & \bfseries airliner     & \bfseries wagon  & \bfseries hum-bird & \bfseries Siam-cat  & \bfseries ox & \bfseries golden \newline retriever & \bfseries frog & \bfseries zebra & \bfseries Con-ship & \bfseries  truck \\
     \midrule
 & \multicolumn{10}{c}{\bfseries ResNet18}  \\
 -- &\textbf{0.96}	&\textbf{ 0.7}	&\textbf{0.62	}	&\textbf{0.7}	&0.24		&\textbf{0.74}	&\textbf{0.72}	&0.34		&\textbf{0.82}	&0.4  \\
 HFSS &\textbf{1	}	&0.14		&\textbf{0.78	}	&\textbf{0.64}		&0.14		&0.22		&0.18		&0.1		&0.32		&0.14	 \\
 \cmidrule{2-11}
  & \multicolumn{10}{c}{\bfseries ResNet34} \\
  --&\textbf{0.96}& \textbf{0.72}&\textbf{0.64	}&	\textbf{0.82}&	0.24	&\textbf{0.66}&	\textbf{0.68}&0.2	&	\textbf{0.84}&	0.48	\\
      HFSS &\textbf{1}		&0.08		&\textbf{0.9	}	&\textbf{0.68	}	&0.44		&0.46		&0.28		&0		&0.18		&0.32	\\
    \cmidrule{2-11}
     & \multicolumn{10}{c}{\bfseries ResNet50} \\
     --& \textbf{0.96}&\textbf{0.76}	&\textbf{0.6}	&\textbf{0.78}&0.28	&\textbf{0.74}&\textbf{0.74}	&0.22	&\textbf{0.84}&0.44	\\
      HFSS& \textbf{0.98}		&0.28		&\textbf{0.86}		&0.48		&0.26		&0.26	&0.32	&0.22		&0.28		& 0.32	\\
    \bottomrule
  \end{tabular}
\end{table}

%\newpage
\subsection{C-10}
We report the test results of ResNet models trained on C-10 in~\cref{tab:idin10}.  All models trained on C-10 learn shortcuts to classify images in classes \textit{airplane} and \textit{bird}. Based on the threshold value of TPR (0.6), ResNet34 and ResNet50 are less subject to frequency shortcuts, although they still learn them, indicating that larger model capacity is not sufficient to avoid shortcut learning, in line with~\cite{Wang_2023_ICCV}.

The OOD test results of C-10 models are provided in~\cref{tab:oodcifar}. Models exhibit close-to or above-average TPR for classes \textit{airliner} and \textit{humming bird}, which is attributable to the presence of shortcuts in the OOD data.

\begin{table}[!t]
\tiny  
  \caption{$\mathrm{TPR}$ results of ResNet50 on DFM-filtered IN-10 images. $\mathrm{TPR}{\geq}0.6$ (a strong frequency shortcut) is highlighted in bold.}
  \label{tab:compareSI_rn50}
  \centering
  \renewcommand{\arraystretch}{0.85}
  \begin{tabular}{p{1.1cm}@{\hspace{1\tabcolsep}}p{2.5em} @{\hspace{1\tabcolsep}}p{2.5em} @{\hspace{1\tabcolsep}}p{2.5em} @{\hspace{1\tabcolsep}}p{2.5em} @{\hspace{1\tabcolsep}}p{3.em} @{\hspace{1\tabcolsep}}p{3.em} @{\hspace{1\tabcolsep}}p{2.5em} @{\hspace{1\tabcolsep}}p{2.5em} @{\hspace{1\tabcolsep}}p{2.5em} @{\hspace{1\tabcolsep}}p{2.5em}@{\hspace{1\tabcolsep}}}
 
     \toprule %\ \newline Method
     \bfseries   \diagbox[innerwidth=0.9cm,height=0.5cm]{Method}{Class}     & \bfseries airliner     & \bfseries wagon  & \bfseries hum-bird & \bfseries Siam-cat  & \bfseries ox & \bfseries golden \newline retriever & \bfseries frog & \bfseries zebra & \bfseries Con-ship & \bfseries  truck \\
    \midrule
 
       HFSS  & \textbf{1	}	& 0.06	& 0.3&\textbf{0.94	}&\textbf{0.76	}	&0.46	&0.46	&	\textbf{0.78	}&\textbf{0.82	}	&0.08	\\
     ~\cite{Wang_2023_ICCV} & 0.54  & 0 & 0 & 0.42  & 0 & 0.2  & 0 & 0.16  & \textbf{0.7}  & 0.1\\

    \bottomrule
  \end{tabular}
  \vspace{-1em}
\end{table}

\subsection{IN-10}
We  provide the comparison of the results of ResNet50 using DFMs  searched by HFSS and the algorithm in~\cite{Wang_2023_ICCV} in~\cref{tab:compareSI_rn50}. 
ResNet50 learns strong shortcuts for classes \textit{airliner}, \textit{siamese cat}, \textit{ox}, \textit{zebra} and \textit{container ship}. Although it has larger model capacity than ResNet18, HFSS confirms that it still exploits shortcuts for many classes, in line with the observation in~\cite{Wang_2023_ICCV}. By comparing the TPR values on IN-10 images processed by DFMs obtained through our HFSS algorithm and through that in~\cite{Wang_2023_ICCV}, our algorithm is more effective at finding shortcuts (more TPRs are highlighted in bold).

\begin{table} 
\tiny 
  \caption{Models trained on  IN-10  are  tested on IN-SCT. TPRs higher than or close to average TPR (0.374) are highlighted in bold. }
  \label{tab:oodinsct_rn50}
  \centering
  \renewcommand{\arraystretch}{0.85}
  \begin{tabular}{p{1.1cm}@{\hspace{1\tabcolsep}}p{2.5em} @{\hspace{1\tabcolsep}}p{2.5em} @{\hspace{1\tabcolsep}}p{2.5em} @{\hspace{1\tabcolsep}}p{2.5em} @{\hspace{1\tabcolsep}}p{3.em} @{\hspace{1\tabcolsep}}p{3.em} @{\hspace{1\tabcolsep}}p{2.5em} @{\hspace{1\tabcolsep}}p{2.5em} @{\hspace{1\tabcolsep}}p{2.5em} @{\hspace{1\tabcolsep}}p{2.5em}@{\hspace{1\tabcolsep}}}
 
     \toprule
     \bfseries  \diagbox[innerwidth=0.9
     cm,height=0.5cm]{Method}{Class}   &  \bfseries Mil-aircraft     & \bfseries car  & \bfseries lorikeet & \bfseries tabby cat & \bfseries holstein & \bfseries Lab-retriever  & \bfseries  tree frog & \bfseries horse & \bfseries fishing vessel & \bfseries fire truck   \\
    \midrule
      -- & \textbf{0.429} &\textbf{ 0.486} & \textbf{0.414} & 0.2    & \textbf{0.37} & 0.3    & 0.3    & 0.057 & \textbf{0.44} & \textbf{0.743} \\
     \bfseries HFSS  & 0.257		& \textbf{0.514}	& \textbf{0.5}		& 0.3	& \textbf{0.372	}	& \textbf{0.386	}	&\textbf{ 0.371	}	& 0		& \textbf{0.486}		& \textbf{0.7}	 \\
     \cite{Wang_2023_ICCV}  &  0.243 & 0      & 0.057 & 0.043 & 0      & 0.2    & 0      & 0      &\textbf{0.486} & 0.043 \\      
    \bottomrule
  \end{tabular}
\end{table}

We report the TPRs of ResNet50 tested on IN-SCT in~\cref{tab:oodinsct_rn50}. The model achieves  higher or close to average TPR of classes \textit{holstein} and \textit{fishing vessel} in IN-SCT, which is attributable to the shortcuts for classifying classes \textit{ox} and \textit{container ship} in IN-10.  Comparing the TPR values of HFSS and ~\cite{Wang_2023_ICCV}, we observe that weak shortcuts for some classes  e.g. \textit{frog} and \textit{golden retriever} are still present in the OOD data, but~\cite{Wang_2023_ICCV} fails to recognize them, demonstrating the effectiveness of HFSS in finding shortcuts. 
 
\subsection{Visualization of images filtered by DFMs obtained over five trials}
\begin{figure} 
    \centering
    \subfloat[]{\includegraphics[width=1.0\linewidth]{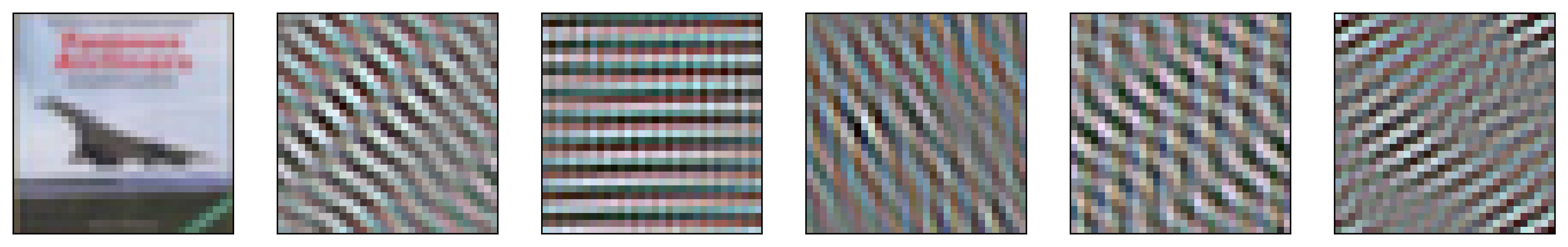} }
    
    \subfloat[]{\includegraphics[width=1.0\linewidth]{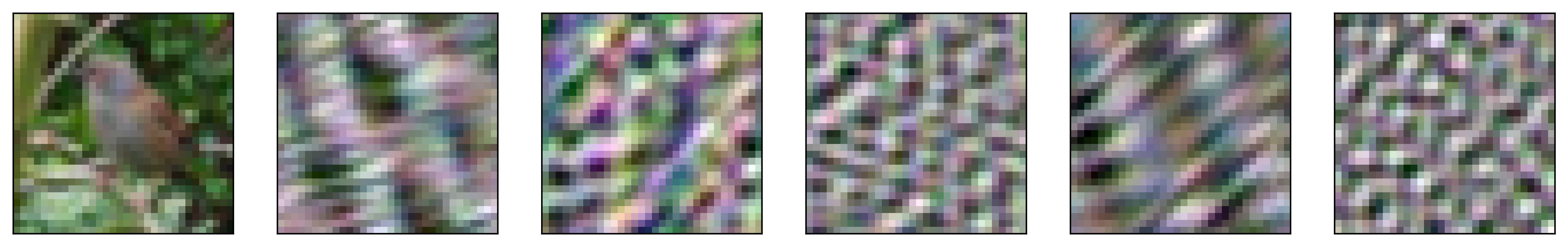} }

    \subfloat[]{\includegraphics[width=1.0\linewidth]{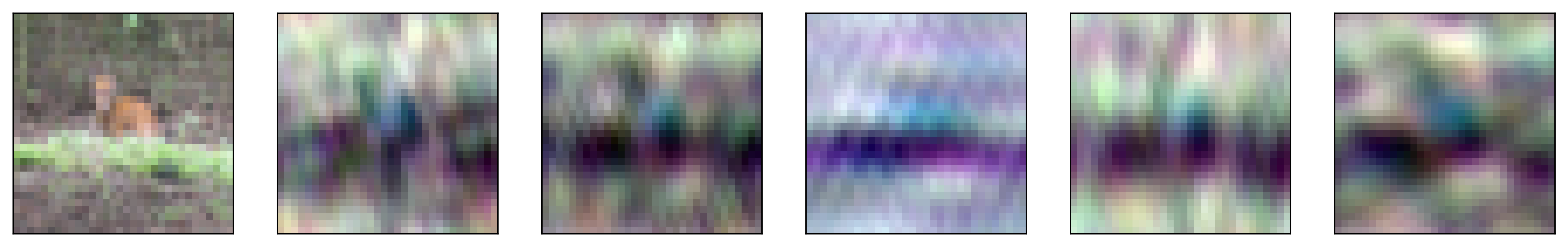} }

    \subfloat[]{\includegraphics[width=1.0\linewidth]{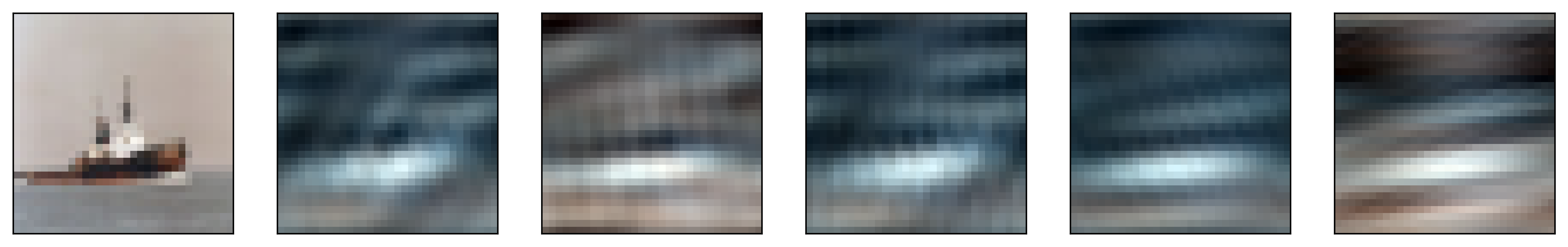} }
    
    \caption{Images of classes (a)  \textit{airplane}, (b)   \textit{bird}, (c)   \textit{deer} and (d) ship in C-10 filtered by corresponding DFM obtained through five trials. We   normalize the images to a range of 0 to 1 for visualization purpose. }
    \label{fig:5trials}
\end{figure}

Due to random sampling of candidate frequency subsets, the outcomes of HFSS might deviate slightly for each run. However, from the visualization of the  image filtered by DFM obtained over five trials, we observe similar texture shortcuts (see~\cref{fig:5trials}). This indicates that frequency shortcuts are not formed by a fixed set of frequencies, but correspond to similar spatial patterns. 

\end{document}